%% file: main.tex
\pdfoutput=1
\documentclass[11pt]{article}

\usepackage[final]{acl}

\usepackage{times}
\usepackage{dsfont}
\usepackage{latexsym}
\usepackage{microtype}
\usepackage{amssymb}
\usepackage{tcolorbox}
\usepackage{amsmath}
\usepackage{fdsymbol}
\usepackage{bbm}
\usepackage{inconsolata}
\usepackage{float}
\usepackage{multirow}
\usepackage{graphicx}
\usepackage{booktabs}
\usepackage{caption}
\usepackage{subcaption}
\usepackage{bbold}
\usepackage{xspace}
\usepackage{makecell}
\usepackage{enumerate}
\usepackage{booktabs}
\usepackage[capitalize,noabbrev]{cleveref}
\usepackage{tcolorbox}
\usepackage[T1]{fontenc}

\usepackage[utf8]{inputenc}

\newcommand\eg{\emph{e.g.},\xspace}

\newcommand\aka{\emph{a.k.a.}\xspace}
\newcommand\ie{\emph{i.e.},\xspace}

\newcommand{\papertitle}{\textsc{GoldCoin}\xspace}
\newcommand{\datasettitle}{\textsc{GoldCoin-hipaa}\xspace}

\newcommand{\textttbf}[1]{\texttt{\textbf{#1}}\xspace}
\definecolor{stepcolor}{HTML}{d79b00}
\definecolor{contentcolor}{HTML}{6c8ebf}

\definecolor{cYellow}{RGB}{255,255,3}
\definecolor{cBlue}{RGB}{69,123,157}
\definecolor{cRed}{RGB}{231,56,71}
\definecolor{cRed_1}{RGB}{191,30,46}
\definecolor{cGray}{RGB}{168,218,219}
\definecolor{cBlue_2}{RGB}{5,48,97}
\definecolor{cBlue_1}{RGB}{115,186,214}
\definecolor{cBlue_3}{RGB}{13,76,109}
\definecolor{cBlue_4}{RGB}{64,121,160}
\definecolor{cOrange}{RGB}{250,134,0}
\definecolor{cBlue_6}{RGB}{13,76,109}
\definecolor{cBlue_7}{RGB}{16,106,130}
\definecolor{cBlue_8}{RGB}{19,136,160}
\definecolor{cBlue_9}{RGB}{115,184,214}
\title{\papertitle: Grounding Large Language Models in Privacy Laws via Contextual Integrity Theory}

\author{
Wei Fan\thanks{\quad Equal Contribution},
Haoran Li$^{*}$\thanks{\quad Corresponding Author},
Zheye Deng,
Weiqi Wang,
Yangqiu Song\\
Department of Computer Science and Engineering, HKUST, Hong Kong SAR, China\\
\texttt{\{wfanag, hlibt, zdengah\}@connect.ust.hk}, \texttt{\{wwangbw, yqsong\}@cse.ust.hk}\\
}

\begin{document}
\maketitle
\begin{abstract}
Privacy issues arise prominently during the inappropriate transmission of information between entities.
Existing research primarily studies privacy by exploring various privacy attacks, defenses, and evaluations within narrowly predefined patterns, while neglecting that privacy is not an isolated, context-free concept limited to traditionally sensitive data~(\eg social security numbers), but intertwined with intricate social contexts that complicate the identification and analysis of potential privacy violations.
The advent of Large Language Models~(LLMs) offers unprecedented opportunities for incorporating the nuanced scenarios outlined in privacy laws to tackle these complex privacy issues.
However, the scarcity of open-source relevant case studies restricts the efficiency of LLMs in aligning with specific legal statutes. 
To address this challenge, we introduce a novel framework, \papertitle\footnote{The code and data are available at \url{https://github.com/HKUST-KnowComp/GoldCoin}}, designed to efficiently ground LLMs in privacy laws for judicial assessing privacy violations.
Our framework leverages the theory of \textit{contextual integrity} as a bridge, creating numerous synthetic scenarios grounded in relevant privacy statutes (e.g., HIPAA), to assist LLMs in comprehending the complex contexts for identifying privacy risks in the real world.
Extensive experimental results demonstrate that \papertitle markedly enhances LLMs' capabilities in recognizing privacy risks across real court cases, surpassing the baselines on different judicial tasks.
\end{abstract}

\input{latex/1_introduction}
\input{latex/2_related_work}

\input{latex/3_method}
\input{latex/4_dataset}
\input{latex/5_experiment}
\input{latex/6_conclusion}
\input{latex/limitation_ethics}

\section*{Acknowledgements}
The authors of this paper were supported by the NSFC Fund (U20B2053) from the NSFC of China, the RIF (R6020-19 and R6021-20), and the GRF (16211520 and 16205322) from RGC of Hong Kong.

\newpage
\bibliography{custom}
\input{latex/appendix}

\end{document}

%% file: latex/1_introduction.tex
\section{Introduction}
\begin{figure}[t!]
    \centering
    \includegraphics[width=0.42\textwidth]{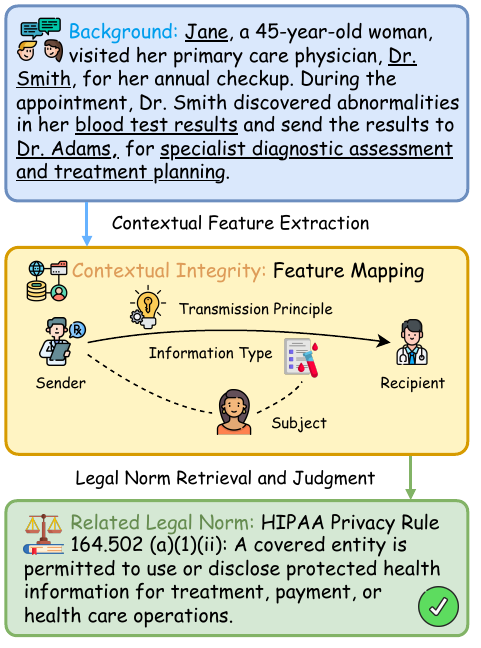}
    \caption{An overview of how our proposed \papertitle bridges the case background and legal norm through contextual integrity theory~\cite{nissenbaum2004privacy}.}
    \label{figs:introduction}
\end{figure}

Privacy violations happen through improper information transmission, including the disclosure of personally identifiable information, inappropriate data collection, and unauthorized access, all of which contradict societal expectations~\cite{martin2016measuring} and legal statutes such as HIPAA~\cite{act1996health}, COPPA~\cite{aftab1999children}, and GDPR~\cite{voigt2017eu}.
In the past few decades, current research has mainly focused on exploring privacy violations in limited pre-defined patterns or manually annotated rules, such as RBAC~\cite{sandhu1998role, kuhn2010adding}, EPAL~\cite{ashley2003enterprise, ashley2002p3p}, thereby diminishing the capacity to detecting privacy risks across diverse social contexts.

Intuitively, we consider applying the wealth of real-world scenarios contained in legal statutes and case law to address the limitation.
However, converting legislation into an actionable framework remains a significant challenge.
Previous efforts have involved translating legislation into logical languages \cite{lam2009formalization,deyoung2010experiences,robaldo2020formalizing}, yet this method heavily relies on expert annotation and struggles to adapt to legislative changes or scale across different privacy laws.
The recent emergence of LLMs~\cite{openai2022chatgpt, touvron2023llama, anthropic2024introducing} has introduced new potential for addressing the problem.
Specifically, legal LLMs like LawGPT~\cite{zhou2024lawgpt}, Lawyer LLaMA~\cite{huang2023lawyer}, ChatLaw~\cite{cui2023chatlaw} have all leveraged the vast existing statutes and cases to assist public in general legal tasks.

Nonetheless, aligning LLMs with specific privacy laws is a non-trivial task. 
The scarcity of open-source public court cases makes it challenging to ensure that the datasets used in model training are comprehensive enough to encompass all aspects of the laws. 
This limitation significantly undermines the LLMs' ability to generalize to unfamiliar cases. 
Moreover, we observe unstable and limited improvements when training LLMs directly on statutory laws~(\cref{sec:overall_performance}), as court cases generally provide a richer source of practice-oriented information, such as factual backgrounds, judicial analyses, and judge opinions.


To fill these gaps, we introduce \papertitle, a novel framework that \underline{\textbf{G}}r\underline{\textbf{O}}unds \underline{\textbf{L}}arge Language Mo\underline{\textbf{D}}els into Privacy Laws via \underline{\textbf{CO}}ntextual \underline{\textbf{IN}}tegrity, which is a theory proposed by~\citet{nissenbaum2004privacy} to assess the appropriateness of privacy information flows. 
Within contextual integrity, privacy information flows are conceptualized as activities involving three relevant entities: the sender, the recipient, and the subject of the information. 
It argues that entities do not merely act as individuals in an undifferentiated social world~\cite{barth2006privacy}, but rather as individuals playing various roles within specific contexts~(\eg healthcare, education, employment). 
Within each distinct context, information flows are regulated by norms~(\aka regulations, legal clauses) that specify the types of the transmitted information and the transmission principles~(\eg purpose, consent, belief).
Then we can abstract privacy laws as the framework for determining the legality of information flow in diverse contexts, including entities, information type, and transmission principles.
Each clause in privacy laws, such as \texttt{164.502(a)(1)(ii)} referenced in~\cref{figs:introduction}, can be interpreted as a legal-grounded norm, either permitting or forbidding information transmission.

Based on this, \papertitle combines the formalization of contextual integrity with concrete seed norms in privacy laws to generate the synthetic background stories by GPT-4~\cite{achiam2023gpt4}. 
To ensure high-quality generation, we employ automatic filters to select cases that include essential features~(\eg sender, recipient) in contextual integrity and are consistent with the seed norms.
Additionally, we develop a diversity ranking mechanism to improve the semantic diversity of the case backgrounds, enhancing training robustness.
Ultimately, our framework combines background contexts and seed norms to construct synthetic court cases tailored to specific privacy laws.

For evaluation, we develop the case dataset \datasettitle under the HIPAA Privacy Rule~\cite{act1996health}, including a ground-truth benchmark sourced from the Caselaw Access Project~(CAP)\footnote{\url{https://case.law/}}~\cite{chang2020mining}, which collectes numerous real-world court cases in the United States.
We experiment with several transformer-based LLMs by instruction-tuning them with~\papertitle.
The evaluation results demonstrate that our synthetic dataset effectively aids LLMs in comprehending privacy laws. 
The models tuned with our framework show superior ability in identifying the applicability of HIPAA in real cases, surpassing other baselines by 8\% to 23\%. 
Meanwhile, these models show enhanced capabilities in detecting privacy risks, outperforming others by 8\% to 18\%.
Moreover, human analysis and ablation studies confirm the efficacy of contextual integrity in case synthesis and the enhancements in data quality provided by the automatic filter and diversity ranking.


%% file: latex/2_related_work.tex
\section{Related Work}
\subsection{Privacy and Contextual Integrity}
To effectively ground language models into privacy laws for judgment in reality,
we first introduce the contextual integrity theory~\cite{nissenbaum2004privacy} and propose a brief framework based on the existing works~\cite{barth2006privacy, lam2009formalization}.

\paragraph{Roles, Information, and Transmission Principle}
Each information transmission inherently involves three main entities $\mathcal{P}$: the \textit{\textbf{sender}}, \textit{\textbf{recipient}}, and \textit{\textbf{subject}} whose information is about. 
The roles of these entities are deeply contextual, as individuals participate in specific \textit{\textbf{roles}} $\mathcal{R}$ tailored to distinct social contexts such as healthcare and commerce. 
Moreover, the \textit{\textbf{information type}} associated with the subject is denoted as $\mathcal{T}$. \textit{\textbf{Transmission principles}}, represented as $\Omega$, comprise specific constraints $\omega \in \Omega$~(\eg purpose, authorization) that regulate the information flow.

\paragraph{Expressing in Norms}
Applying contextual integrity to the privacy law $\mathcal{L}$, we can abstract each legal clause as a norm $n$, which governs the information flow between entities.
\begin{equation} \label{alg:norms}
\begin{aligned}
    &\textit{permitted}_{by}n^{+} \iff (\mathcal{P},\mathcal{R}) \land \mathcal{T} \land \Omega,\\
    &\textit{forbidden}_{by}n^{-} \iff (\mathcal{P},\mathcal{R}) \land \mathcal{T} \land \Omega,\\
\end{aligned}
\end{equation}
where a \textbf{\texttt{permit}} norm $n^{+}$ allows an information transmission when satisfying conditions, and a \textbf{\texttt{forbid}} norm $n^{-}$ prohibits it when aligning with the specified features. 
Further details and examples are provided in the~\cref{app:contextual_integrity}.

\subsection{LLMs in Law}
Recent advancements in legal LLMs, such as LawGPT~\cite{nguyen2023brief, zhou2024lawgpt}, Lawyer LLaMA~\cite{huang2023lawyer, Touvron2023LLaMAOA} and SaulLm~\cite{colombo2024saullm7b}  have shown significant improvements in a broad array of legal services, including judgment prediction~\cite{yue2021neurjudge, 10.1145/3580489}, court view generation~\cite{10.1145/3404835.3462984}, and question answering~\cite{Duan_2019, zhong2020jec}.
ChatLaw~\cite{cui2023chatlaw} specializes in processing Chinese legal queries, excelling in keyword extraction and court case similarity-matching. 
However, these LLMs often underperform in the privacy domain, where related training and evaluation datasets are limited and generally close-sourced.

\subsection{LLMs for Instruction Generation}
A series of works have explored the use of LLMs for data generation~\cite{10.5555/3618408.3619426, Liu2022WANLIWA, schick-schutze-2021-generating,DBLP:conf/acl/0001FLS0XWBLJCS24,DBLP:journals/corr/abs-2406-02106}. 
Recent studies have particularly concentrated on enhancing instruction generation~\cite{honovich2022unnatural, zhou2023large, singh2023explaining, Honovich2022InstructionIF, wang-etal-2023-self-instruct} to improve zero-shot~\cite{Ye2022GuessTI} and few-shot~\cite{DBLP:journals/corr/abs-2005-14165, DBLP:journals/corr/abs-2109-01652} learning, abstraction reasoning~\cite{wang2024absinstruct} capabilities, as well as the instruction-following proficiency of LLMs. 
Inspired by it, our approach utilizes the strong generative capabilities of GPT-4 to address the case scarcity based on contextual integrity theory by instructing it to generate datasets and align LLMs with privacy laws for judicial.

%% file: latex/3_method.tex
\section{Method}
\begin{figure}[t]
    \centering
    \includegraphics[width=0.9\columnwidth]{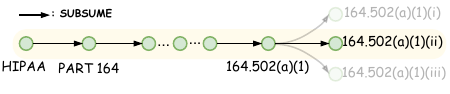}
    \vspace{-0.1in}
    \caption{We concatenate all the content along the whole path from the leaf~(\texttt{164.502(a)(1)(ii)}) to the root~(\texttt{HIPAA}) node and refer to it as a norm, as illustrated in the norm part of~\cref{tabs:synthetic_case_permit_example}.}
    \vspace{-0.25in}
    \label{figs:norm}
\end{figure}

\begin{figure*}[t]
    \centering
    \includegraphics[width=2\columnwidth]{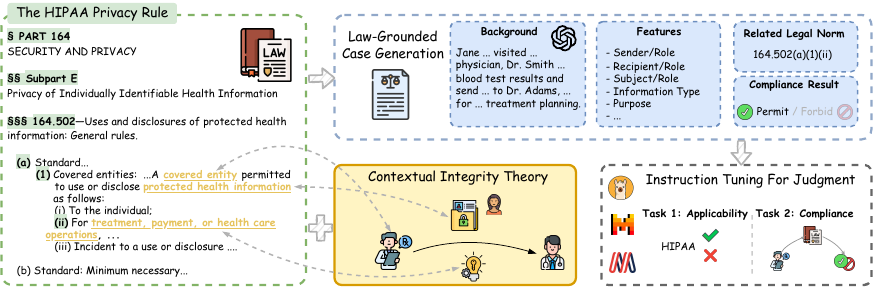}
    \caption{The overview of our \papertitle framework. We use \texttt{164.502(a)(1)(ii)} as a seed norm to generate cases based on the contextual integrity theory and instruction-tune the models for downstream judicial tasks.}
    \label{figs:method}
\end{figure*}

Legal judgments on privacy violations typically involve two tasks~\cite{lam2009formalization}: (1) \textbf{Applicability}, assessing whether the privacy law $\mathcal{L}$ applies to the case background $s$; and (2) \textbf{Compliance}, determining if the transmission described in $s$ compliant with $\mathcal{L}$.
In this section, we introduce \papertitle, which applies contextual integrity theory to generate synthetic cases. After postprocessing the instances, we instruction-tune LLMs and evaluate their performance in the above two tasks.
The overview of our pipeline is shown in~\cref{figs:method}.




\subsection{Legal Statute Preprocessing}
\label{sec:preprocessing}
To evaluate the effectiveness of \papertitle, we apply it to the U.S. Health Insurance Portability and Accountability Act (HIPAA) Privacy Rule. Initially, we dump the content of the HIPAA Privacy Rule from the official Code of Federal Regulations (CFR) website\footnote{\url{https://www.ecfr.gov/current/title-45/subtitle-A/subchapter-C}}. 
We then transform the textual data into a structured graph $\mathcal{G}$, comprising nodes $\mathcal{V}$ that represent sections and two types of relations $\mathcal{E}$. These relationships are identified as \textbf{\texttt{subsume}}, denoting hierarchical relationships (e.g., (\texttt{164.502(a)}, \texttt{subsume}, \texttt{164.502)}), and \textbf{\texttt{refer}}, indicating cross-references between sections (e.g., (\texttt{164.502(a)(1)(ii)}, \texttt{refer}, \texttt{164.504(b)})). 
Each node consists of a labeled identifier and the paragraph content. We start from each leaf node $v^l_i$ and recursively identify all parent nodes $\{v^{l-1}_i, v^{l-2}_i,\ldots, v^0_i\}$ where $v^0_i$ is the root node. Then, we aggregate their content, as depicted in~\cref{figs:norm}, and refer to each such path as a norm.
These norms can be categorized into three types: \textbf{\texttt{permit}}~$n^{+}$, \textbf{\texttt{forbid}}~$n^{-}$, and others.
The first two categories describe the permissions and prohibitions regarding information transmission under the law, while the last category contains general definitions, exceptions, and requirements.
We leverage GPT-4 to classify and label these leaf norms and filter the permit and forbid norm for the subsequent generation steps. The examples of permit and forbid norms are shown in the upper part of~\cref{tabs:synthetic_case_permit_example} and~\cref{tabs:synthetic_case_forbid_example}, respectively. All the details of the HIPAA Privacy Rule and preprocessing are depicted in~\cref{app:HIPAA}.

\subsection{Law-Grounded Case Generation}\label{sec:case_generation}
After classification, we select the norms $\mathcal{N} = \{n_1, n_2, \ldots, n_m\}$, a filtered subset of $\mathcal{L}$, as seeds for case synthesis. Our objective is to generate the case set $\mathcal{K} = \{k_1, k_2, \ldots, k_m\}$, with each case $k_i$ derived from $n_i$.
In a synthetic case, four key elements are considered: \textit{case background}, \textit{contextual features}, \textit{related norm}, and \textit{conclusion}.

\paragraph{Instruction Compilation with Norm}
Given the seed norm $n_i$ and the conclusion $c_i$, which correspond to the norm type (\ie permit, forbid), we manually build the instructions combined with $n_i$ for background generation. To ensure the generation of background narratives that align with $n_i$ and preserve the integrity of the privacy information transmission context, we construct a detailed prompt~(see~\cref{app:case_generation_prompt}) that includes the description of the key features in contextual integrity, such as entities, roles, information type, and transmission principles.

\paragraph{Response Collection and Parsing}
To enhance the reliability of the model outputs, we sample several responses for each norm. Following the collection of GPT-4 outputs, we parse the responses and focus on the five components of $k_i$: (1) \textbf{\textit{Background}} $s_i$, which is the background description of the information transmission. (2) \textbf{\textit{Contextual features}} $\{(\mathcal{P}_i,\mathcal{R}_i), \mathcal{T}_i, \Omega_i\}$, which denotes the key features in the transmission context. (3) \textbf{\textit{Norm}} $n_i$, which denotes the related legal clause (\eg \texttt{164.502(a)(1)(ii)}) to the generated background, (4) \textbf{\textit{Applicability conclusion}} $c^{appl}_i$, which denotes whether the case applies to $\mathcal{L}$. (5) \textbf{\textit{Compliance conclusion}} $c^{comp}_i$, which represents whether $n_i$ permits or forbids the case.

\subsection{Case Postprocessing}
After collecting all cases, we implement several filters to ensure the consistency and quality of the selected cases.

\paragraph{Contextual Feature Integrity Filter}
By analyzing the key characteristics that are entailed in the generated cases, we observe that GPT-4 sometimes omits some main features in contextual integrity due to unstable instruction-following ability. To ensure the integrity of context in case background, we filter out all cases that lack any vital features in \textbf{\textit{sender, sender role, recipient, recipient role, subject, subject role,}} and \textbf{\textit{information type}}.

\paragraph{Consistency Filter}
Each synthetic case is derived from a specific norm; however, the probabilistic variability of GPT-4 outputs may result in the related norm $\hat{n}$ of the synthetic case not aligning with the initial seed norm $n$.
To improve the consistency of the cases, we filter out the cases that are not related to the given seed norm:
\begin{equation} \label{algs:filter_norm}
    f_\text{norm}(n, \hat{n}) = \mathds{1}(n =\hat{n}),
\end{equation}
where $f_\text{norm}$ denotes the compare function between the seed norm and the case norm.
Moreover, we expect the model to generate cases applicable to $\mathcal{L}$ and its compliance $c^{comp}$ (\ie permit or forbid) is consistent with the type of seed norm~$n^{+/-}$:
\begin{equation} \label{algs:filter_compliance_applicable}
\begin{aligned}
    &f_\text{conc}(c^{appl}) =\mathds{1}(c^{appl} = applicable),\\
    &f_\text{conc}(n^{+/-}, c^{comp}) = \mathds{1}(n^{+/-}=c^{comp}),
\end{aligned}
\end{equation}
Then $f_\text{conc}$ filters out all conclusion-inconsistent cases, ensuring that all cases apply to $\mathcal{L}$ and compliance with the seed norms.

\paragraph{Diversity Ranking}
To mitigate the reduction in semantic complexity and alignment robustness caused by similar features across different case backgrounds, we implement the methodology from prior studies~\cite{wang-etal-2023-self-instruct, wang2024absinstruct} to promote background diversity. We calculate the ROUGE-L~\cite{lin-2004-rouge} similarity for each case background against others, ranking them accordingly. For each norm, we select the case with the highest ROUGE-L score to ensure optimal diversity.

\begin{figure}[!t]
    \centering
    \includegraphics[width=0.8\columnwidth]{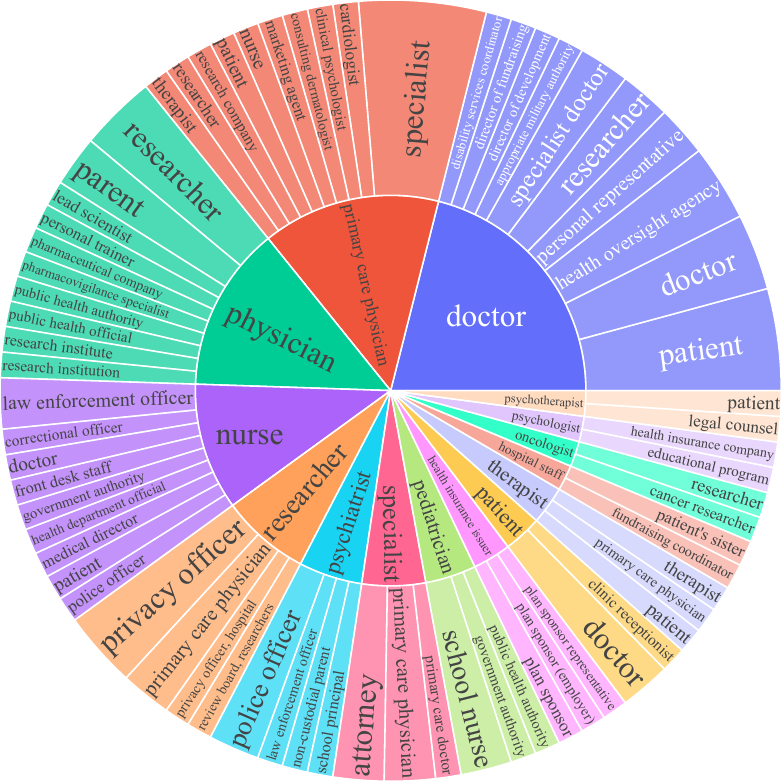}
    \caption{Top 15 common sender roles (inner circle) and their top 10 recipient roles (outer circle).}
    \label{figs:sender_recipient}
\end{figure}

\subsection{Real Court Case Collection}\label{sec:real_case_collection}
To rigorously validate the grounding efficacy of \papertitle, we retrieve all relevant real court cases featuring the ``HIPAA Privacy Rule" from the Caselaw Access Project~(CAP)\footnote{\url{https://case.law/}}, developed by Harvard Law School. These cases are systematically processed and distilled into the same format as synthetic cases, utilizing both GPT-4 and manual annotation.
Additionally, as delineated in~\cref{sec:case_generation}, our methodology focuses exclusively on constructing cases applicable to the HIPAA Privacy Rule because privacy cases unrelated to HIPAA are readily available. To provide a negative training and testing set for the applicability task, we also curate a collection of cases that, while closely related to privacy violations, do not apply to HIPAA. The details of CAP are explained in~\cref{app:CAP}.

\definecolor{deepblue}{RGB}{110,144,193}
\definecolor{babyblue}{RGB}{218,232,252}
\subsection{Instruction and Response Compilation}
Using the case background as input, we manually build multi-step instructions (see~\cref{tabs:applicability_instruction}(b) and~\cref{tabs:compliance_instruction}(b)) with the Chain-of-Thought~(CoT) prompting~\cite{DBLP:journals/corr/abs-2201-11903}, for both applicability and compliance tasks as follows:

\begin{center}
\vspace{-10pt}
\resizebox{1\linewidth}{!}{
\begin{tabular}{l}
\noindent\textttbf{\textcolor{stepcolor}{Instruction: }\textcolor{contentcolor}{<task-specific instruction>}}\\
\noindent\textttbf{\textcolor{stepcolor}{Input: }\textcolor{contentcolor}{<case background>}}\\
\end{tabular}
}
\end{center}

We construct responses for the applicability task following two steps: first, extracting context-specific features; second, determining HIPAA applicability:

\begin{center}
\begin{tabular}{l}
\noindent\textttbf{\textcolor{stepcolor}{Step1: }\textcolor{contentcolor}{<sender>}, \textcolor{contentcolor}{<recipient>}, ...}\\
\noindent\textttbf{\textcolor{stepcolor}{Step2: }\textcolor{contentcolor}{Applicable/Not applicable}}
\end{tabular}
\end{center}

In the compliance task, we initially guide LLMs to extract features based on contextual integrity, focusing on principles of information transmission. Subsequently, we retrieve the relevant norm and finally determine whether the norm permits or prohibits the transmission as the format:


\begin{center}
\begin{tabular}{l}
\noindent\textttbf{\textcolor{stepcolor}{Step1: }\textcolor{contentcolor}{<sender>}, \textcolor{contentcolor}{<recipient>}, ...}\\
\noindent\textttbf{\textcolor{stepcolor}{Step2: }\textcolor{contentcolor}{<norm id>}, \textcolor{contentcolor}{<norm content>}}\\
\noindent\textttbf{\textcolor{stepcolor}{Step3: }\textcolor{contentcolor}{Permit/Forbid}}
\end{tabular}
\end{center}





%% file: latex/4_dataset.tex
\section{\datasettitle Dataset Overview}
In this section, we apply our framework to the HIPAA Privacy Rule as a case study and introduce the \datasettitle dataset. Statistics of the dataset are provided in Table~\ref{tabs:dataset}.

\subsection{\papertitle Synthetic Cases}
\begin{figure}[!t]
    \centering
    \includegraphics[width=1\columnwidth]{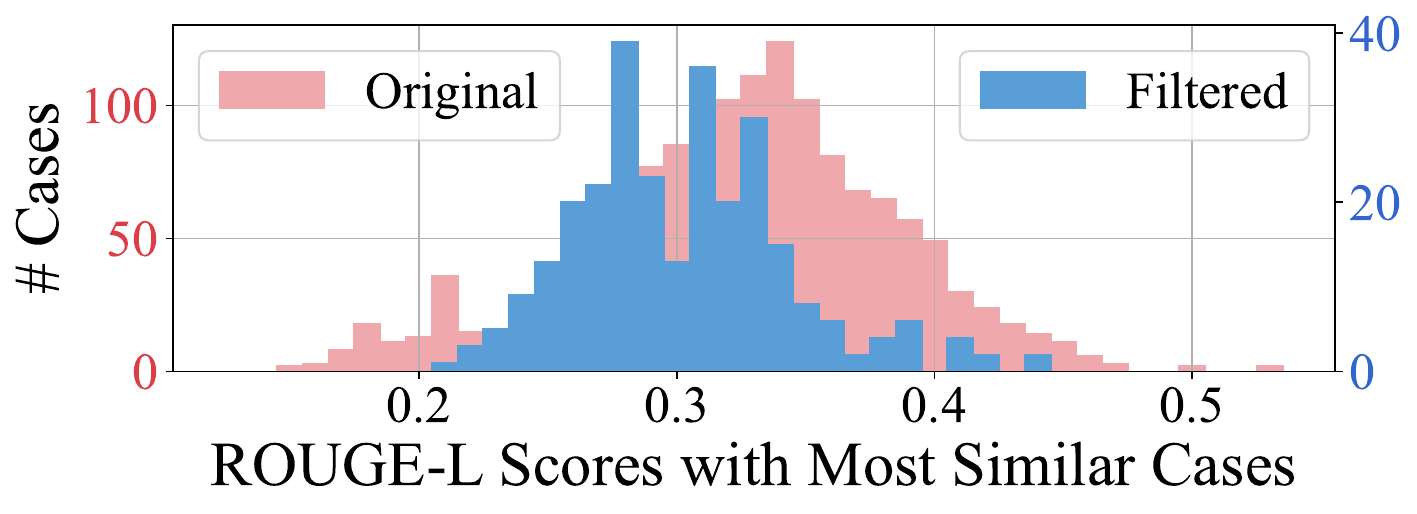}
    \caption{The ROUGE-L score distribution between the original and filtered cases.}
    \label{figs:diversity}
\end{figure}

In HIPAA, we analyze 269 \texttt{permit} and 40 \texttt{forbid} norms, generating multiple cases for each norm. We filter and select the case with the highest ROUGE-L scores for each norm, ultimately collecting 309 cases.
Comparisons between original and filtered cases in terms of ROUGE-L score distributions are shown in~\cref{figs:diversity}, indicating the efficacy of diversity ranking.
Moreover, we plot a sunburst chart to display the top 15 most common roles of \textit{\textbf{sender}} and their top 10 \textit{\textbf{recipients}} (Figure~\ref{figs:sender_recipient}), and another chart to detail the common combination between information subjects and types (Figure~\ref{figs:subject_type.pdf}).
\paragraph{Human Analysis}
To further investigate the quality, we enlist two experts to evaluate the synthetic cases using the criteria outlined in Table~\ref{tab:data_quality_eval}. The results confirm that all cases apply to HIPAA, with the majority being related to the seed norm.

\begin{table}[t]
\small
\setlength\doublerulesep{\arrayrulewidth}
 \renewcommand\cellset{\renewcommand\arraystretch{1}}
\centering
\begin{tabular}{lc}
\toprule
Quality Review Question & Yes \%\\ 
\midrule
\midrule
\makecell[l]{Does HIPAA apply to this case?}&100.0\%  \\ \midrule
\makecell[l]{Is the case strongly related to the seed norm?}&99.35\%  \\ \midrule
\makecell[l]{Is the compliance of the case correct?}&99.03\%  \\ \midrule\midrule
All fields are valid&  98.38\%\\ 
\bottomrule
\end{tabular}
\caption{Human analysis of synthetic case quality.}
\label{tab:data_quality_eval}
\end{table}

\subsection{CAP Real Court Cases}
Due to reasons outlined in Section~\ref{sec:real_case_collection}, we directly collect cases irrelevant to HIPAA as negative training examples. We select cases tagged with ``Privacy Violation'' using the ``Most Relevant First'' search function on the CAP website, gathering 309 non-applicable cases for training.
For evaluation, after a combined screening by GPT-4 and human experts, we identify 107 real court cases relevant to HIPAA, serving as the ground truth for the compliance task. Correspondingly, we also sample an equivalent number of HIPAA-irrelevant cases and combine them with the 107 cases to form the test set for the applicability task. Ultimately, we combine synthetic and real cases to create the \datasettitle dataset. \cref{app:CAP} details the collection and post-processing of CAP cases.

%% file: latex/5_experiment.tex
\section{Experiment}\label{sec:experiment}
In this section, we conduct extensive experiments to demonstrate the efficacy of \papertitle in grounding LLMs into real-world privacy laws.

\input{tabs/evaluation_overall_performance}

\subsection{Experimental Settings}
\paragraph{Datasets and Metrics}
As illustrated in Table~\ref{tabs:dataset}, our framework generates 309 synthetic cases that either comply with or violate the HIPAA Privacy Rule~\cite{act1996health}. Also, we collect 309 cases that do not apply to HIPAA. For evaluation, we collect 107 HIPAA-related and 107 unrelated real court cases from the CAP and calculate Accuracy~(Acc) and Macro F1-score~(Ma-F1) as metrics between predicted and ground truth conclusion.

\paragraph{Models}
We conduct instruction tuning on four open-source LLMs: MPT-7B-Chat-8k~\cite{MosaicML2023Introducing}, Mistral-7B-Instruct-v0.2~\cite{jiang2023mistral}, Llama-2-7b-chat-hf and Llama-2-13b-chat-hf~\cite{touvron2023llama}. These models all support at least 4k tokens content length and have superior instruction-following ability.
Additionally, we evaluate our method against closed-source LLMs in zero-shot and few-shot settings, including models such as ChatGPT~(gpt-3.5-turbo)~\cite{openai2022chatgpt} and GPT-4~(gpt-4)~\cite{achiam2023gpt4,openai2024gpt4}, both with the version \texttt{2024-02-01} via Azure OpenAI API.

\paragraph{Baseline Methods}
We conduct comparative experiments against the following baselines to demonstrate the improvement introduced by \papertitle.
(1)~Zero-shot: Given the background of cases, the LLMs should directly determine whether the case applies to HIPAA and violates HIPAA or not.
(2)~Law Recitation: No learning from cases, we tune the LLMs directly on the legal norm content.
(3)~Direct Prompt: Different from zero-shot, we instruction-tune the LLMs with vanilla prompts, where the responses are solely (``Applicable,'' ``Not Applicable'') or (``Permit,'' ``Forbid'').
The baseline prompts are shown in~\cref{app:implement_details}.

\subsection{Overall Performance}\label{sec:overall_performance}
\begin{figure}[!t]
    \centering
    \includegraphics[width=1\columnwidth]{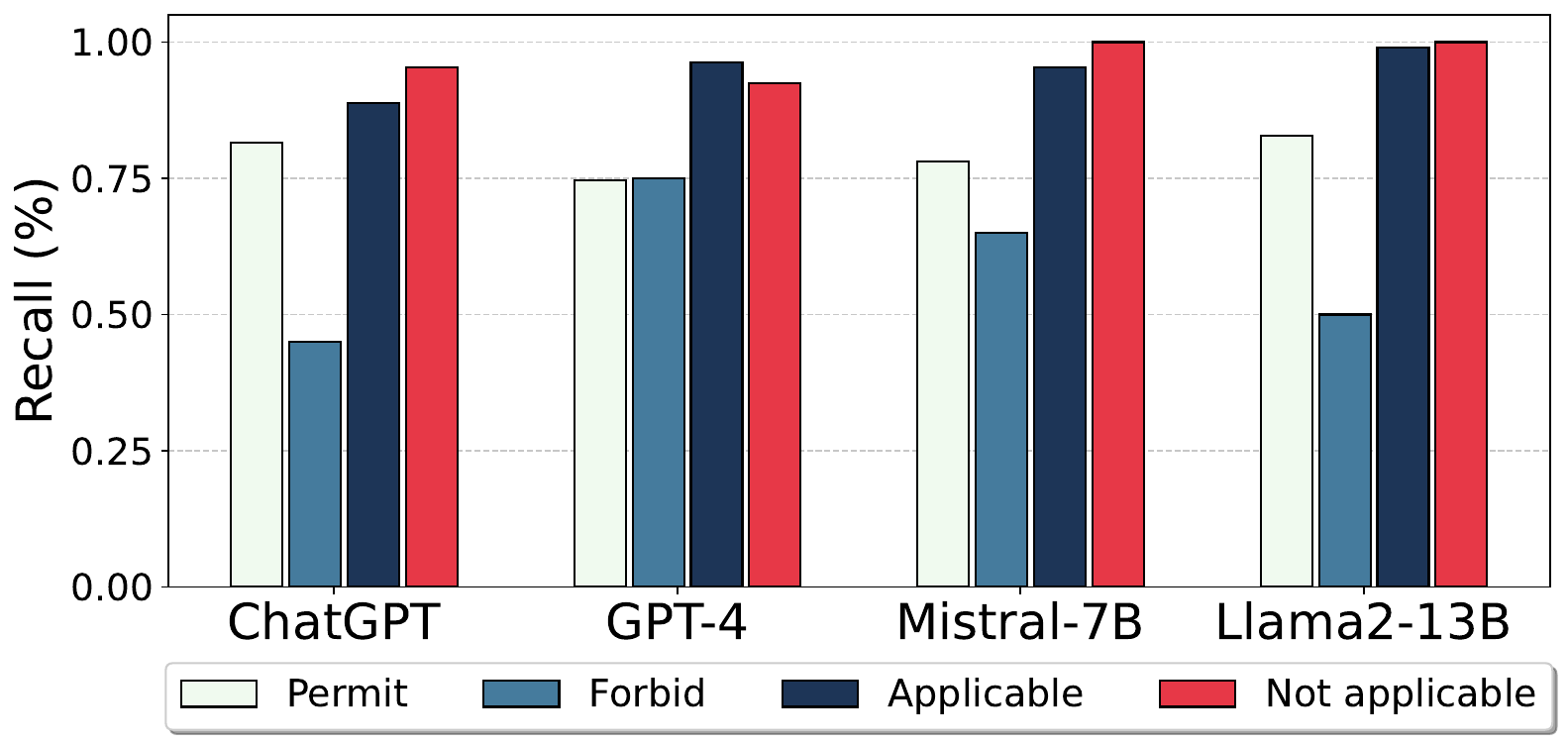}
    \caption{Comparative performance of GPT series models and our GoldCoin framework measured by \textbf{Recall} across all categories, with multi-step instructions.}
    \vspace{-0.15in}
    \label{figs:gpt_compare}
\end{figure}

We present comprehensive results for two judicial tasks in Table~\ref{tabs:overall_performance}, which includes the baseline methods and our \papertitle. Besides, Figure~\ref{figs:gpt_compare} displays a comparison results with the GPT-series.
\paragraph{Applicability}
We first analyze the performance of four LLMs in determining the HIPAA applicability of real court cases sourced from CAP.
Our results demonstrate that \papertitle can align the LLMs with the comprehensive understanding of the HIPAA Privacy Rule, exceeding all baseline methods. Notably, MPT-7B, which performed near-random levels~(Acc ~50\%, Ma-F1 ~50\%), see substantial improvements with our method—accuracy and Macro F1-scores increase by 12.62\% and 11.81\%, respectively, compared to the zero-shot setting.
Meanwhile, Mistral-7B and Llama2-13B,  tuned with our framework, achieve exceptional accuracy rates of 97.66\% and 99.53\%, respectively, even attaining 100\% in the ``Not applicable'' category~(see~\cref{tabs:full_overall_performance_applicability}), surpassing the performance of ChatGPT and GPT-4.
We observe that MPT-7B, when trained exclusively with ``Direct Prompt,'' exhibits only a limited improvement of 2.49\% in Ma-F1. This underscores the integration of contextual features, which is crucial for decomposing and deeply understanding legal case topics. Additionally, our results indicate that merely continuing to train LLMs on legal statutes results in limited effectiveness and even leads to diminished performance in determining applicability~(\eg MPT-7B $\downarrow$10.8\%).

\paragraph{Compliance}
Our \papertitle framework introduces multi-step simulated trial instructions, effectively aligning LLMs with privacy law and enhancing their reasoning capabilities on compliance tasks. It significantly improved Macro F1-scores across several models: MPT-7B~(17.87\%), Llama2-7B~(12.45\%), Mistral-7B~(17.96\%), and Llama2-13B~(8.12\%) compared to the zero-shot setting.
Mistral-7B, specifically tuned on our dataset, excels in precision for both ``permit'' and ``forbid'' cases, surpassing ChatGPT and approaching GPT-4's performance. However, using ``Direct Prompt'' results in a notable decline for Mistral-7B, from 66.98\% to 51.75\%, indicating limited grounding ability. Direct training on abstract legal concepts leads to reasoning confusion, as seen with Llama2-7B, which tends to misclassify cases as ``forbid'' (see~\cref{tabs:full_overall_performance_compliance}).
Our results reaffirm the high quality of cases generated under contextual integrity theory and the feasibility of the reasoning pipeline for adjudicating privacy law cases.
Furthermore, to demonstrate that sample imbalance does not affect overall results, we apply oversampling to the forbid cases and assess the performance impact on the Mistral-7B and Llama2-13B models. Our findings indicate no significant change in the final performance, as detailed in \cref{tabs:oversampling} in \cref{app:oversampling}.

\input{tabs/ablation_llama13b_mistral}
\subsection{Ablation Study}
To better understand how to ensure the quality of synthetic cases grounded in real law, we conduct several ablation studies. These studies demonstrate the effectiveness of our contextual feature filter, consistency checks, and diversity ranking. The complete results of these ablation studies are presented in~\cref{tabs:ablation_all}.

\paragraph{Contextual Feature Filter}
We conduct ablation studies to assess the effect of contextual feature filters. After generating case backgrounds, we retain all cases, including those that lacked key features~(\eg \textit{sender}, \textit{recipient}) of contextual integrity. The results, denoted as ($\diamond$ w/o Feature F), reveal significant performance declines. Specifically, there is a drop of 3.26\% and 2.36\% in the applicability and compliance tasks, respectively, for Llama2-13B~(see~\cref{tabs:ablation_llama13b_mistral}). These findings demonstrate the importance of feature integrity.

\paragraph{Consistency Filter}
First, we remove the norm consistency filter ($\diamond$ w/o Norm F) and do not verify whether the legal norms in synthetic cases match the seed norms. Here, Mistral-7B drops by 3.62\% in the compliance task illustrating the efficacy of the norm consistency checker in mitigating issues such as hallucinations during generation.
Subsequently, we observe a significant performance decline when we bypass the check of the conclusion ($\diamond$ w/o Conclusion F). Incorrect conclusions lead to increased perplexity in legal judgments during training, which in turn causes a 4.99\% drop in the applicability judgments for Llama2-13B.

\paragraph{Diversity Ranking}
We remove the diversity ranking ($\diamond$ w/o Diversity R) and randomly sample cases for each norm. Low diversity often results in high similarity among cases, such as in the roles of entities or specific categories of information. The lack of diversity can decrease the robustness of training, as demonstrated in \cite{wang2024absinstruct, wang-etal-2023-self-instruct}. This impact is further reflected in a 3.86\% decline in the Macro F1-score for applicability judgments in Llama2-13B. Furthermore, we deactivate all of the above filters and ranking mechanisms ($\diamond$ w/o All Parts) and observe significant decreases across all language models, with Mistral-7B experiencing drops of 5.89\% and 5.05\% in each task, respectively. These findings underscore the importance of enhancing the integrity, consistency, and diversity of generated cases.

\subsection{Discussion of \papertitle Instruction}
\input{tabs/step_evaluation}
To further investigate whether the improvement in model performance stems from the quality of synthetic cases or the instructions themselves, we conduct experiments utilizing multi-step instructions on all baseline models~(see results in~\cref{tabs:overall_performance_cot}). Additionally, we discuss how contextual integrity affects norm retrieval accuracy and judgment performance as shown in~\cref{tabs:step_evaluation}.

\paragraph{Multi-step Instruction}\label{sec:multi-step}
As shown in \cref{tabs:overall_performance_cot}, we can compare this with \cref{tabs:overall_performance} and notice that the Macro F1-scores for MPT-7B and Mistral-7B exhibit a slight average improvement of 1.70\% when determining the applicability of HIPAA under the zero-shot setting.
Nonetheless, the Llama2 series shows a decline of 2.17\%, indicating unstable performance when not aligned with specific cases.
Similar results are reflected in the compliance tasks, demonstrating that merely relying on detailed instructions is insufficient to guide LLMs to follow contextual integrity for effective judgment. The instability may arise when the models are not exposed to such case types and legislation during pre-training, underscoring the importance of our approach that utilizes synthetic cases grounded in actual laws.
 
\paragraph{Features in Contextual Integrity}
Contextual Integrity~(CI)~\cite{nissenbaum2004privacy} serves as a bridge between abstract privacy laws and specific cases, enhancing norm retrieval and subsequently improving judgment capabilities.
We omit the contextual feature extraction step in the compliance task (w/o CI), and the results are presented in~\cref{tabs:step_evaluation}. The norm retrieval accuracy declines significantly across all open-source LLMs tuned by \papertitle, demonstrating that contextual features effectively aid the model in understanding information transmission within cases and aligning them with pertinent legal statutes.
Llama2-13B, which exhibits the best norm retrieval performance, experiences a significant decrease of 5.14\% in conclusion performance when contextual integrity features are not extracted. 
These findings substantiate that contextual integrity is an effective formalization method in the privacy domain, further demonstrating the efficacy of our \papertitle framework in aligning LLMs with privacy laws.

%% file: tabs/evaluation_overall_performance.tex
\begin{table*}[!t]

    \renewcommand\arraystretch{1.06}
    \small
    \begin{center}           
    \begin{tabular}{m{1.9cm}|m{1.9cm}|m{0.95cm}<{\centering}m{0.95cm}<{\centering}|m{0.95cm}<{\centering}m{0.95cm}<{\centering}|m{0.95cm}<{\centering}m{0.95cm}<{\centering}||m{0.95cm}<{\centering}m{0.95cm}<{\centering}}
            \toprule
            \multirow{2}{*}{\textbf{Task}}&\multirow{2}{*}{\textbf{Method}} & \multicolumn{2}{c|}{\textbf{MPT-7B}} &  \multicolumn{2}{c|}{\textbf{Llama2-7B}}  & \multicolumn{2}{c|}{\textbf{Mistral-7B}} &\multicolumn{2}{c}{\textbf{Llama2-13B}}\\
            && {Acc} & {Ma-F1} & {Acc} &  {Ma-F1} & {Acc} &  {Ma-F1} & {Acc} &  {Ma-F1} \\
            \midrule
            \multirow{4}{*}{{\textbf{Applicability}}}
            &Zero-shot       &55.61&55.49&72.89&71.05&89.25&89.24&91.12&91.07\\ 
            &Law Recitation  &44.86&44.69&74.30&72.75&85.98&85.96&91.59&91.57\\
            &Direct Prompt   &63.55&57.97&89.25&89.13&95.33&95.32&94.39&94.39\\
            &\textbf{\papertitle} &68.22&67.30&94.39&94.39&\underline{97.66}&\underline{97.66}&\textbf{99.53}&\textbf{99.53}\\
            
            \midrule
            \multirow{4}{*}{{\textbf{Compliance}}}
            &Zero-shot       &46.73&40.75&56.07&47.14&50.47&49.02&65.42&56.71\\ 
            &Law Recitation  &39.25&32.43&42.99&41.69&53.27&43.23&68.22&59.79\\
            &Direct Prompt   &66.36&56.46&62.62&53.68&53.27&51.75&73.83&62.40\\
            &\textbf{\papertitle} &69.16&58.62&\textbf{79.44}&59.58&75.70&\textbf{66.98}&\underline{76.64}&\underline{64.83}\\
            \bottomrule
    \end{tabular}
    \end{center}
    \vspace{-0.1in}
    \caption{Performance of four LLMs under three baselines and our \papertitle, showing \textbf{Acc} and \textbf{Ma-F1} across both applicability and compliance tasks. We \textbf{bold} the best results and \underline{underline} the second-best results in each task.}
    \vspace{-0.15in}
    \label{tabs:overall_performance}
\end{table*}

%% file: tabs/ablation_llama13b_mistral.tex
\begin{table}[t]
\setlength{\tabcolsep}{3pt}
    \small
	\centering
	\begin{tabular}{l|cccc}
	\toprule
        \multirow{1}{*}{\textbf{Model}}&\multicolumn{1}{c}{\textbf{\scriptsize{Applicability}}} &\multicolumn{1}{c}{\textbf{$\Delta$\textsubscript{App}}}&\multicolumn{1}{c}{\textbf{\scriptsize{Compliance}}}&\multicolumn{1}{c}{\textbf{$\Delta$\textsubscript{Com}}}\\
            \midrule
            Llama2-13B &\textbf{99.53}&-&\textbf{64.83}&-\\
            \midrule
            $\diamond$ w/o Feature F &96.27&$\downarrow$3.26&62.47&$\downarrow$2.36 \\
            $\diamond$ w/o Norm F &97.59&$\downarrow$1.94&61.34&$\downarrow$3.49\\
            $\diamond$ w/o Conclusion F &94.54&$\downarrow$4.99&61.07&$\downarrow$3.76 \\
            $\diamond$ w/o Diversity R &95.67&$\downarrow$3.86&62.33&$\downarrow$2.50 \\
            \midrule
            $\diamond$ w/o All Parts &93.01&$\downarrow$6.52&60.11&$\downarrow$4.72\\
            \midrule
            \midrule
            Mistral-7B &\textbf{97.66}&-&\textbf{66.98}&-\\
            \midrule
            $\diamond$ w/o Feature F &95.22&$\downarrow$2.44&65.04&$\downarrow$1.92 \\
            $\diamond$ w/o Norm F &95.98&$\downarrow$1.68&63.34&$\downarrow$3.62\\
            $\diamond$ w/o Conclusion F &93.61&$\downarrow$4.05&63.05&$\downarrow$3.91 \\
            $\diamond$ w/o Diversity R &95.54&$\downarrow$2.12&64.45&$\downarrow$2.51 \\
            \midrule
            $\diamond$ w/o All Parts &91.77&$\downarrow$5.89&61.91&$\downarrow$5.05\\
		\bottomrule
	\end{tabular}
	\caption{Ablation study for \papertitle. Macro F1-scores are presented, with \textbf{$\Delta$} indicating score changes.}
        \vspace{-0.15in}
    \label{tabs:ablation_llama13b_mistral}
\end{table}

%% file: tabs/step_evaluation.tex


\begin{table}[t]
    \renewcommand\arraystretch{1.06}
    \small
    \centering
    \setlength{\tabcolsep}{5pt}
    \begin{tabular}{l|cc|cc}
    \toprule
       \multirow{2}{*}{\textbf{Models}} &\multicolumn{2}{c|}{$\textbf{Norm.}_{Acc}$}&\multicolumn{2}{c}{$\textbf{Conc.}_{Ma-F1}$}\\
       &{w/ CI}&w/o CI&{w/ CI}&w/o CI\\
    \midrule
    {MPT-7B}  &34.58&29.91&58.62&53.44  \\
    {Llama2-7B}  &46.73&39.25&59.58&56.72  \\
    {Mistral-7B}    &51.40&45.79&\underline{66.98}&61.22  \\
    {Llama2-13B}  &\underline{53.27}&43.93&64.83&59.69  \\
    \bottomrule
    \end{tabular}
    \caption{Performance comparison with and without contextual feature extraction in the first step during tuning and evaluation. $\textbf{Norm.}_{Acc}$ denotes norm retrieval accuracy, and $\textbf{Conc.}_{Ma-F1}$ indicates Macro F1-scores of conclusions~(permit, forbid).}
    \label{tabs:step_evaluation}
\end{table}

%% file: latex/6_conclusion.tex
\section{Conclusion}
In this paper, we introduce \papertitle, a pioneering framework that leverages the contextual integrity theory to effectively apply privacy laws to privacy violation detection.
Specifically, we practice the HIPAA Privacy Rule and build synthetic cases for aligning LLMs.
Our experimental results demonstrate that this approach significantly enhances models' capability to assess legal relevance and pinpoint privacy risks, providing a novel perspective for the integration of privacy legislation within LLMs.
In the future, this generation and alignment method could be extended to other privacy laws such as GDPR and COPPA, or general legal domains. We hope our \papertitle sheds light on the development of legal LLMs.

%% file: latex/limitation_ethics.tex
\section*{Limitations}
Our methodology rigorously adheres to the \texttt{permit} and \texttt{forbid} norms delineated within HIPAA; however, it fails to incorporate the interconnections among these norms.
Legal norms frequently entail cross-references, wherein adjudication of cases may hinge upon multiple norms simultaneously, as evidenced by the formalization of legal reasoning~\cite{Eisenberg_2022,lam2009formalization}.
Future work should construct cases based on multiple norms to reflect real-world scenarios better and potentially yield improvements.
Additionally, we do not consider the few-shot setting due to multiple examples often exceeding the maximum input length of LLMs.
For the selection of laws, we conduct experiments on HIPAA due to its prominence in the privacy domain and the relatively abundant availability of open-source cases, which can serve as ground truth for testing.
We invite legal professionals with access to cases related to other privacy laws to contact us, as this would facilitate the extension of our approach to additional privacy regulations such as COPPA~\cite{aftab1999children}, GDPR~\cite{voigt2017eu}, etc.
Moreover, this paper primarily focuses on case generation, and we do not employ techniques such as retrieval-augmented generation~\cite{gao2024retrievalaugmented, DBLP:journals/corr/abs-2005-11401} or vector embedding~\cite{douze2024faiss} for retrieving relevant norms. We believe that dynamically indexing~\cite{Liu_LlamaIndex_2022} and retrieving related norms, based on the statute graph constructed in~\cref{sec:preprocessing}, is a promising direction.

\section*{Ethics Statement}
We use the API provided by the official website of the Code of Federal Regulations to access the HIPAA Privacy Rule.
We follow contextual integrity theory~\cite{nissenbaum2004privacy} to generate 
synthetic cases for constructing \datasettitle, and manually remove cases that could potentially leak real-world private information.
We follow the official usage and access rules of the Caselaw Access Project\footnote{\url{https://case.law/about/\#usage-access}} during downloading relevant cases.
Human evaluations and annotations are performed by two legal experts to review the quality of synthetic cases and remove cases that contain explicit content such as gore or violence.
Annotations are compensated at 15 USD per hour, above the local minimum wage.
To the best of our knowledge, this work complies with open-source agreements and does not pose risks of information leakage or other hazards.


%% file: latex/appendix.tex
\newpage

\appendix

\section{Contextual Integrity: Theory and Framework}
\label{app:contextual_integrity}

In this appendix, we explore the concept of contextual integrity as developed by~\citet{nissenbaum2004privacy}. This theory serves as a framework~\cite{barth2006privacy} for formalizing information transmission, particularly within various societal contexts.

\subsection{Information Transmission}
Information transmission involves three primary entities: the \textit{\textbf{sender}} of the message, the \textit{\textbf{recipient}} who receives the information, and the \textit{\textbf{subject}} who is related to the information, also referred to as the \textit{\textbf{about}}.
The \textit{\textbf{information type}} \( t \in \mathcal{T} \) is another crucial element, referring to the specific category of the transmitted information~(\eg health plan, address).
These elements constitute the fundamental components of transmission.

\subsection{Roles and Contexts}
\label{app:contexts}
At the core of contextual integrity lies the concept of \textbf{context}. Nissenbaum emphasizes that individuals operate not merely as undifferentiated entities but in specific roles within different social contexts, such as healthcare, education, employment, and marketplaces. Each entity within a context plays specific roles \( r \in \mathcal{R} \).
Understanding these roles is crucial as they significantly influence the nuanced judgments individuals make concerning potential privacy violations. For instance, Mr. Smith, depicted in~\cref{figs:introduction}, may act as a doctor within a healthcare setting, subject to HIPAA~\cite{act1996health}, a consumer in a supermarket, subject to the CCPA~\cite{pardau2018california}, or a father within his family setting. Each role carries distinct expectations and norms regarding privacy. Accurately identifying and comprehending the role of entities within the specific context is essential for determining the appropriate law to apply in privacy risk detection.

\subsection{Transmission Principles}
After understanding the concept of information flows and context, we then expand to the concept of \textit{\textbf{transmission principle}}, which is a distinctive aspect of the contextual integrity approach to privacy. These principles define the specific constraints regulating the flow of information from one entity to another. In this work, we select \textbf{\textit{Purpose, In Reply To, Consented By, Belief}} as the key transmission principles. The meanings of these principles are shown in~\cref{app:case_generation_prompt}.
Future extensions to other privacy legislations could involve adding new principles manually or guiding LLMs to automatically induce principles based on the target laws.

\subsection{Informational Norms}
With all features of contextual integrity in place, we introduce the concept of \textit{\textbf{norm}}. Norms governing the transmission of personal information from one party to another, referred to as ``informational norms'', are derived from societal expectations and legal standards. These norms restrict, for example, what physicians can disclose about the health conditions of patients under their care. Since societal expectations are challenging to define and subjective, this work relies on standardized legal frameworks to extract norms. We can represent a norm of information flow as $(\mathcal{P},\mathcal{R}) \land \mathcal{T} \land \Omega$.
Legal regulations such as HIPAA provide a formal definition for each type of information transmission, as expressed abstractly in~\cref{alg:norms}. Then the legality of each information transmission can be defined as:

\begin{equation}
\begin{aligned}
&\operatorname{inrole}(p_{s}, \hat{r}_{s}) \wedge \operatorname{inrole}(p_{r}, \hat{r}_{r}) \wedge \operatorname{inrole}(p_{a}, \hat{r}_{a})\\
&\wedge (t \in \hat{t}) \wedge \Omega \rightarrow \{\text{permit, forbid}\},\\
\end{aligned}
\label{algs:detail_norm}
\end{equation}
where \( p_{s}, p_{r}, p_{a} \) represent the sender, recipient, and subject, respectively.
Besides, current research~\cite{confaide2023} explores expressing personal privacy expectations as norms to assess privacy risks. This approach also represents a valuable area for further exploration.

\subsection{Example}
To illustrate the application of norms, we consider the example from~\cref{figs:introduction}. With the theory of contextual integrity, we map the features of the healthcare context to the formal representation in~\cref{algs:detail_norm} as follows:
\begin{equation}
\begin{aligned}
&\operatorname{inrole}(p_{s}, \text{doctor}) \wedge \operatorname{inrole}(p_{r}, \text{doctor}) \wedge\\
&\operatorname{inrole}(p_{a}, \text{patient}) \wedge
(t \in \text{blood test results}) \wedge\\
&(\omega_{purp} \in \text{treatment planning}),
\end{aligned}
\label{algs:example_norm}
\end{equation}
where \( \omega_{purp} \) denotes the purpose of the information transmission.
Given that the doctor is a covered entity under HIPAA and blood test results are health information, this information transmission aligns with the legal norm \texttt{164.502(a)(1)(ii)} of HIPAA, thereby being permitted.

\section{HIPAA Privacy Rule} \label{app:HIPAA}

\subsection{Brief Introduction}
The Health Insurance Portability and Accountability Act (HIPAA) of 1996 Title II sets national standards for protecting personal health information (PHI). Defined as PHI, this includes any individually identifiable health information managed by entities such as health plans, health care clearinghouses, and health care providers who transmit health information electronically. The HIPAA Privacy Rule, detailed in \texttt{45 CFR Parts 160}, \texttt{162}, and \texttt{164}, provides federal protections for PHI, limits its disclosure without consent, and gives patients rights regarding their health information, such as accessing and amending their records. The Privacy Rule is codified at \texttt{45 CFR Part 160}\footnote{\url{https://www.ecfr.gov/current/title-45/subtitle-A/subchapter-C/part-160?toc=1}} and \texttt{Subparts A and E of Part 164}\footnote{\url{https://www.ecfr.gov/current/title-45/subtitle-A/subchapter-C/part-164}}.

\subsection{Requirement, Exception, and General Definition}
In \cref{sec:preprocessing}, besides norms like \textbf{\texttt{permit}} and \textbf{\texttt{forbid}} that describe compliance with privacy information transmission, HIPAA also includes the following basic types of norms:
\begin{itemize}
    \item \textbf{\texttt{Requirement}} indicates that an action is permissible under the rule only if specific conditions are met. For example, according to \texttt{164.508(a)(2)}, an action is allowable only with proper authorization.
    \vspace{-8pt}
    \item \textbf{\texttt{Exception}} refers to a specific scenario where a standard rule or requirement does not need to be applied. For instance, \texttt{164.508(a)(2)} specifies that if psychotherapy notes are used for the treatment of the originator, the usual authorization requirement is waived.
    \vspace{-8pt}
    \item \textbf{\texttt{General definition}} provides a broad explanation of concepts or terms. For example, in HIPAA terminology, a ``covered entity'' is defined as a health plan, a health care clearinghouse, or a health care provider who transmits health information in electronic form.
\end{itemize}

Single norm may consist of multiple types (e.g., permit with requirement, permit with exception). In this work, we focus only on norms containing the permit and forbid types.

\subsection{Details of Norm Classification}
In this appendix, we provide the prompt for norm classification in~\cref{tabs:norm_classfication_prompt}. We compile each norm with the classification instruction and utilize GPT-4 to extract the basic norm type.

\paragraph{Statistics}

In this work, we mainly focus on the \texttt{45 CFR Part 164}, which governs security and privacy concerns in the healthcare sector. Following the classification in Section~\ref{sec:preprocessing}, we analyze the types within each norm, the statistical results are presented in~\cref{tabs:hipaa_statistics}. Our analysis identified 269 out of 691 HIPAA norms that allow certain information transmissions, while 40 out of 691 norms prohibit specific transmissions. However, the classification of norm types by LLMs is not always accurate. As demonstrated in~\cref{tab:data_quality_eval}, there are three instances where the compliance results conflict with the specified norm type. Following expert annotation, we find out that one permit norm and two forbid norms are misclassified.

\begin{figure}[t]
    \centering
    \includegraphics[width=0.8\columnwidth]{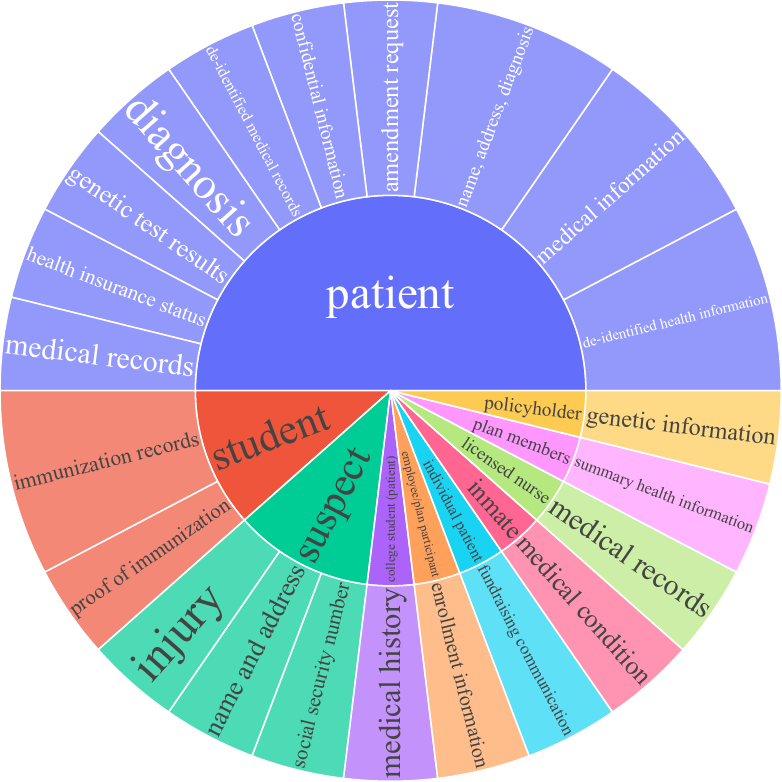}
    \caption{Top 10 common information subjects (inner circle) and their corresponding top 10 information types (outer circle).}
    \label{figs:subject_type.pdf}
\end{figure}

\input{tabs/dataset}

\begin{table}[t]
\small
    \centering
    \begin{tabular}{p{2.5cm}p{1.5cm}}
    \toprule
    \textbf{Norm Type} & \textbf{\# Number} \\
    \midrule
    Total & 691\\
    \midrule
    - Permit & \textbf{269} \\
    - Forbid & \textbf{40} \\
    - Requirement & 555 \\
    - Exception & 112 \\ 
    - Definition & 44 \\ 
    \bottomrule
    \end{tabular}
    \caption{Statistics of norm categories within HIPAA Privacy Rule. Each norm may encompass several norm types, thereby the \textbf{Total} number of norms is less than the cumulative sum of individual types.}
    \label{tabs:hipaa_statistics}
\end{table}

\section{Details of \textsc{GoldCoin-hipaa} Dataset}
\subsection{Prompt of Background Generation}\label{app:case_generation_prompt}
To ensure that contextual integrity is considered when constructing case backgrounds, we incorporate the definition of privacy information flow along with key contextual features into the prompt, as shown in~\cref{tabs:case_generation_prompt}. Based on the provided norm (i.e., regulation, clause) and its type (e.g., \texttt{permit}, \texttt{forbid}), we prompt GPT-4 to generate the corresponding case background.

\subsection{Statistics}
This appendix presents the statistics of the training and testing datasets used in this study. The term ``Synthetic'' refers to cases generated by \papertitle, which are based on HIPAA regulations, while ``Rea'' indicates cases collected from CAP~(see~\cref{app:CAP}) and processed through our pipeline. The statistics of \datasettitle dataset are provided in~\cref{tabs:dataset}.

\subsection{Case Study}
We present two examples to visually show the quality of the cases generated by our framework. The first example is a case permitted under \texttt{164.502(j)(1)(i)}, as shown in~\cref{tabs:synthetic_case_permit_example}. The second example is a case forbidden by the \texttt{164.502(a)(5)(ii)(B)(1)}, as detailed in~\cref{tabs:synthetic_case_forbid_example}. Each case includes one seed norm, a background story, features related to contextual integrity, and a conclusion.

\section{Details of the Caselaw Access Project}\label{app:CAP}
The Caselaw Access Project~(CAP), an initiative by the Harvard Law School Library, has digitized a comprehensive collection of American case law. This monumental effort has converted approximately 40 million pages of court decisions into a machine-readable format, thus making these legal documents accessible online in a consistent format. The collection includes all official, book-published state and federal U.S. case law up to the year 2020, covering a wide range of courts, including state, federal, and territorial courts.

\subsection{Dataset Collection}
We utilized the official API\footnote{\url{https://old.case.law/docs/site_features/api}} provided by CAP, employing ``HIPAA Privacy Rule'' as the keyword for dumping relevant cases.
We filtered out cases longer than 30,000 words and shorter than 100 words before proceeding with further processing.
Additionally, we sampled 2,000 cases related to general privacy violations using the keyword ``privacy violation'' to provide a training and testing set for the applicability task.

\subsection{Prompt}
In this appendix, we provide the prompt as depicted in~\cref{tabs:real_case_process_prompt} for case processing by GPT-4. Since a real case may relate to other legal regulations except HIPAA, we target to extract the factual background, contextual features, related norms, and court conclusions that are relevant to the HIPAA Privacy Rule.

\subsection{Human Annotation}
After the preliminary processing by GPT-4, we engaged two human experts who had studied privacy protection and privacy laws for over a year to manually annotate, correct, and filter the HIPAA-related cases. The annotations focused on three main tasks:(1) Removing cases not related to information transmission. (2) Deleting the court analysis from the background. (3) Assessing whether the court conclusions were correctly extracted.

\subsection{Case Study}
In this appendix, we present three real court cases processed by our pipeline. The first case is an example where HIPAA permits the transmission, as shown in \cref{tabs:real_case_permit_example}. The second case is an example where HIPAA forbids the transmission, as detailed in \cref{tabs:real_case_forbid_example}. The third case is an example that HIPAA is not applicable, as outlined in \cref{tabs:real_case_not_applicable_example}.

\section{Implement Details}\label{app:implement_details}
We select four open-source language models that support at least 4K tokens input and instruction-tune them on one H800 (80G) GPU. Specifically, we parameter-efficient fine-tune MPT-7B, Mistral-7B, Llama2-7B, and Llama2-13B using LoRA. For LoRA, we choose a rank and alpha of 8 and 16, respectively. All language models are trained for 3 epochs, and we select the final checkpoints for evaluation. The batch size is 1, and the learning rate is set to 1e-5.
For API-based LLMs, we access ChatGPT and GPT-4 via the Azure OpenAI API Service\footnote{\url{https://azure.microsoft.com/en-us/products/ai-services/openai-service}}, using the versions \texttt{gpt-3.5-turbo (2024-02-01)} and \texttt{gpt-4 (2024-02-01)}. The total generation and evaluation costs of using the API are approximately \$100 and \$20. respectively.

\subsection{Instruction Template}
To align the knowledge in our case instructions with the language models without compromising its overall performance, we follow the approach described in ~\cite{taori2023alpaca}. The specific prompt used as the instruction template can be found in~\cref{tabs:instruction_template}.

\begin{table}[t!]
\centering
\small
\begin{subtable}[t]{0.96\columnwidth}
\centering
\begin{tabular}{p{0.96\columnwidth}}
\toprule
Below is an instruction that describes a task, paired with an input that provides further context. Write a response that appropriately completes the request.
\newline
\newline
\#\#\# Instruction:
\newline
\textbf{\{instruction\}}
\newline
\newline
\#\#\# Input:
\newline
\textbf{\{input\}}
\newline
\newline
\#\#\# Response: \\
\bottomrule
	\end{tabular}
\caption{Template for examples with an input.}
\end{subtable}

\begin{subtable}[t]{0.96\columnwidth}
\centering
\begin{tabular}{p{0.96\columnwidth}}
\toprule
Below is an instruction that describes a task. Write a response that appropriately completes the request.
\newline
\newline
\#\#\# Instruction:
\newline
\textbf{\{instruction\}}
\newline
\newline
\#\#\# Response: \\
\bottomrule
\end{tabular}
\caption{Template for examples with an empty input.}
\end{subtable}
\caption{The prompt templates used to concatenate instructions and example inputs. Two templates are shown to account for cases where the input is optional. Placeholders \textbf{{instruction}} and \textbf{{input}} are replaced with actual instructions and inputs.}
\label{tabs:instruction_template}
\end{table}


\subsection{Vanilla and Multi-step Prompts for the Applicability Task}
This appendix details the prompts utilized in the applicability task. For the ``Direct Prompt'' along with other baseline approaches, we employed the vanilla prompt, as illustrated in \cref{tabs:applicability_instruction}(a). The response format for the vanilla prompt is straightforward, consisting of either ``Applicable'' or ``Not Applicable''.
Additionally, we implemented multi-step prompts for \papertitle, which are depicted in \cref{tabs:applicability_instruction}(b).

\begin{table}[t]
\small
\centering
\begin{subtable}[t]{0.96\columnwidth}
\centering
\begin{tabular}{p{0.96\columnwidth}}
\toprule
 \textbf{Instruction:} Please determine whether the HIPAA Privacy Rule is applicable to the case.
 \\ \midrule
  \textbf{Input:} Read the case background: <background>.
  \\ \midrule
  \textbf{Response:} Applicable / Not applicable.
 \\ \bottomrule
\end{tabular}
\caption{Vanilla prompt of the applicability task.}
\end{subtable}

\begin{subtable}[t]{0.96\columnwidth}
\centering
\begin{tabular}{p{0.96\columnwidth}}
\toprule
 \textbf{Instruction:} Please assess the applicability of the HIPAA Privacy Rule to the case through the following steps: Step 1: Annotate the message characteristics [Sender, Sender Role, Recipient, Recipient Role, Subject, Subject Role, Type] about the flow of private information in the case as a list. Step 2: Determine whether the HIPAA Privacy Rule is applicable to the case.
 \\ \midrule
  \textbf{Input:} Read the case background: <background>.
  \\ \midrule
  \textbf{Response:} \\
  Step 1: Sender: <sender>, Sender Role: <sender role> ...\\
  Step 2: Applicable / Not applicable.
 \\ \bottomrule
\end{tabular}
\caption{Multi-step instruction of the applicability task.}
\end{subtable}
\caption{The vanilla and multi-step prompts we used in the applicability task.}
\label{tabs:applicability_instruction}
\end{table}

\subsection{Vanilla and Multi-step Prompts for the Compliance Task}

In the compliance task, we utilize the same prompt settings as those in the applicability task. For ``Direct Prompt'', the response to the vanilla prompt is limited to either ``Permit'' or ``Forbid''. This setting is depicted in \cref{tabs:compliance_instruction}(a). For the multi-step instruction presented in \cref{tabs:compliance_instruction}(b), we instruct language models to analyze contextual features and retrieve relevant norms for making informed judgments.

\begin{table}[t]
\small
\centering
\begin{subtable}[t]{0.96\columnwidth}
\centering
\begin{tabular}{p{0.96\columnwidth}}
\toprule
 \textbf{Instruction:} Please determine whether the HIPAA Privacy Rule permits or forbids the case.
 \\ \midrule
  \textbf{Input:} Read the case background: <background>.
  \\ \midrule
  \textbf{Response:} Permit / Forbid.
 \\ \bottomrule
\end{tabular}
\caption{Vanilla prompt of the compliance task.}
\end{subtable}

\begin{subtable}[t]{0.96\columnwidth}
\centering
\begin{tabular}{p{0.96\columnwidth}}
\toprule
 \textbf{Instruction:} Please assess the case for compliance with the HIPAA Privacy Rule through the following steps: Step 1: Annotate the eleven message characteristics [Sender, Sender Role, Recipient, Recipient Role, Subject, Subject Role, Type, Purpose, In Reply To, Consented By, Belief] about the flow of private information in the case as a list. Step 2: Identify and list all applicable HIPAA regulation IDs (e.g., 164.xxx) and their content. Step 3: Determine whether the HIPAA Privacy Rule permits or forbids the case.
 \\ \midrule
  \textbf{Input:} Read the case background: <background>
  \\ \midrule
  \textbf{Response:} \\
  Step 1: Sender: <sender>, Sender Role: <sender role> ...\\
  Step 2: <norm id>, <norm content>.\\
  Step 3: Permit / Forbid.
 \\ \bottomrule
\end{tabular}
\caption{Multi-step instruction of the compliance task.}
\end{subtable}
\caption{The vanilla and multi-step prompts we used in the compliance task.}
\label{tabs:compliance_instruction}
\end{table}

\subsection{Prompt of Law Recitation}
To align language models with the content from privacy laws as a baseline, we build an instruction that guides the models to recite the content of the HIPAA Privacy Rule. As shown in \cref{tabs:law_recitation_instruction}, we incorporate all norms from HIPAA \texttt{Part 164} into this template for tuning.

\begin{table}[t]
\small
\centering
\centering
\begin{tabular}{p{0.96\columnwidth}}
\toprule
 \textbf{Instruction:} Please recite the contents of <norm id> in the HIPAA Privacy Rule.
 \\ \midrule
  \textbf{Response:} <norm content>.
 \\ \bottomrule
\end{tabular}
\caption{Prompt of the baseline ``Law Recitation''.}
\label{tabs:law_recitation_instruction}
\end{table}

\section{Supplementary Experiments}
This section provides additional supplementary experiments to ~\cref{sec:experiment}.

\subsection{Full Results of Ablation Study}
In this appendix, we provide an overall comparison of four LLMs under different ablation settings, focusing on their specific performance deficits as detailed in~\cref{tabs:ablation_all}.
It is observed that inaccuracies in conclusion lead to the most substantial performance degradation, particularly for MPT-7B, which experiences a 6.25\% reduction in accuracy when determining applicability.
The performance loss due to inconsistencies in norms reveals that GPT-4 continues to manifest certain hallucinatory and random behaviors during case generation.

\subsection{Overall Performance across Categories}
In this appendix, we extend the experimental results introduced in~\cref{tabs:overall_performance} across four LLMs and GPT series.
\cref{tabs:full_overall_performance_applicability} provides a comprehensive dataset of experimental results for the applicability task using ``Zero-shot,'' ``Law Recitation,'' ``Direct Prompt,'' and our proposed method \papertitle. Metrics such as Precision (Prec), Recall (Rec), and F1-score (F1) are evaluated for both ``Applicable'' and ``Not Applicable'' categories, alongside average Accuracy (Acc) and Macro F1-score (Ma-F1).
GPT-4 exhibits optimal performance with the ``Direct Prompt'', whereas its efficacy declines when employing multi-step instructions, corroborating the findings discussed in~\cref{sec:multi-step}.
Utilizing \papertitle, Mistral-7B and Llama2-13B achieve 100\% in precision for the positive category and recall in the negative category.
We also provide a detailed analysis of the compliance task as shown in~\cref{tabs:full_overall_performance_compliance} and the inherent instability of the ``Direct Prompt'' is evident; for instance, Mistral-7B reached a precision of 97.44\% in the ``permit'' category, yet the precision for ``forbid'' was merely 27.94\%.
These findings underscore the necessity of integrating our multi-step instructions with the generated cases to achieve optimal outcomes.

\input{tabs/ablation_all}

\subsection{Baselines under Multi-step Instruction}
Table~\ref{tabs:overall_performance_cot} outlines the performances when multi-step instructions are integrated into all baseline models. As discussed in~\cref{sec:multi-step}, the direct application of multi-step prompting in LLMs without instruction-tuning on \datasettitle results in performance degradation. Notably, Llama-2 13B exhibits a 3.33\% decrease in the ``Zero-shot'' setting. This decline is attributed to the model's inability to comprehend and apply contextual integrity without direct reference to legal knowledge.
Furthermore, the top sections of~\cref{tabs:full_overall_performance_compliance} and~\cref{tabs:full_overall_performance_applicability} illustrate how GPT models fare when subjected to multi-step instruction scenarios.

\subsection{Robustness and Sensitivity Analysis}\label{app:oversampling}
Due to the inconsistent quantities of permit and forbid norms in HIPAA, as depicted in~\cref{tabs:hipaa_statistics}, there exists a category imbalance in the generated number of cases.
To ascertain whether the imbalance between norm and case categories in HIPAA adversely influences the efficacy of training and testing, we oversample 269 forbid cases (to match the number of permit cases) and compare the results with the original experiment.
This involved generating ten cases for each of the 40 forbid norms and randomly selecting 269 cases from this pool, striving to maintain an equitable distribution of cases per norm.
The permit type cases remain unchanged as in the uploaded dataset folder.
We select two models, Mistral-7B and Llama2-13B, which perform best in the main experiments. The results, shown in~\cref{tabs:oversampling}, indicate that GoldCoin exhibits insensitivity to sample imbalance, suggesting that the imbalance in training data has a negligible impact on final performance.

\begin{table}[t]
    \centering
    \resizebox{\linewidth}{!}{%
    \begin{tabular}{lccc}
        \toprule
        & \textbf{Permit (F1)} & \textbf{Forbid (F1)} & \textbf{All (Ma-F1)} \\
        \midrule
        Mistral-7B & 83.95 & 50.00 & 66.98 \\
        Mistral-7B$^\spadesuit$ & 84.34 & 45.83 & 65.09 \\
        Llama2-13B & 85.21 & 44.44 & 64.83 \\
        Llama2-13B$^\spadesuit$ & 85.88 & 45.45 & 65.67 \\
        \bottomrule
    \end{tabular}%
    }
    \caption{Comparison of Macro F1-scores for Mistral-7B and Llama2-13B, tuned with and without oversampling data in the compliance task. Models marked with $\spadesuit$ use oversampling during tuning by \papertitle.}
    \label{tabs:oversampling}
\end{table}

\input{tabs/full_overall_performance}
\input{tabs/evluation_overall_performance_cot}

\input{tabs/prompt_norm_classification}
\input{tabs/prompt_case_generation}
\input{tabs/synthetic_case_example}

\input{tabs/not_applicable_case_example}

\input{tabs/prompt_real_case_process}
\input{tabs/real_case_example}

%% file: tabs/dataset.tex
\begin{table}[!t]
\small
    \renewcommand\arraystretch{1.08}
    \centering
    \begin{tabular}{p{3.5cm}p{1cm}<{\centering}p{1cm}<{\centering}}
    \toprule
     $\vardiamondsuit$ \textbf{Applicability} & \textbf{\# Train} & \textbf{\# Test} \\
     \midrule
     Synthetic (Applicable) &  309 & -  \\
     Synthetic (Not Applicable) &  - & -  \\
     Real (Applicable)  &  - & 107  \\
     Real (Not Applicable)  &  309 & 107  \\
     \midrule
     \midrule
     $\vardiamondsuit$ \textbf{Compliance} & \textbf{\# Train} & \textbf{\# Test} \\
     \midrule
     Synthetic (Permit) &  269 & -  \\
     Synthetic (Forbid) &  40 & -  \\
     Real (Permit)  &  - & 80  \\
     Real (Forbid)  &  - & 27  \\
     \bottomrule
    \end{tabular}
    \caption{Statistics of \datasettitle.}
    \label{tabs:dataset}
\end{table}

%% file: tabs/ablation_all.tex
\begin{table}[t]
\setlength{\tabcolsep}{3pt}
    \small
	\centering
	\begin{tabular}{l|cccc}
	\toprule
        \multirow{1}{*}{\textbf{Model}}&\multicolumn{1}{c}{\textbf{\scriptsize{Applicability}}} &\multicolumn{1}{c}{\textbf{$\Delta$\textsubscript{App}}}&\multicolumn{1}{c}{\textbf{\scriptsize{Compliance}}}&\multicolumn{1}{c}{\textbf{$\Delta$\textsubscript{Com}}}\\
            \midrule
            MPT-7B &\textbf{67.30}&-&\textbf{58.62}&-\\
            \midrule
            $\diamond$ w/o Feature F &65.39&1.91$\downarrow$&57.87&0.75$\downarrow$ \\
            $\diamond$ w/o Norm F &66.28&1.02$\downarrow$&55.43&3.19$\downarrow$ \\
            $\diamond$ w/o Conclusion F &61.05&6.25$\downarrow$&53.75&4.87$\downarrow$ \\
            $\diamond$ w/o Diversity R &64.28&3.02$\downarrow$&56.27&2.35$\downarrow$ \\
            \midrule
            $\diamond$ w/o All Parts &60.48&6.82$\downarrow$&53.14&5.48$\downarrow$ \\
            \midrule
            \midrule
            Llama2-7B &\textbf{94.39}&-&\textbf{59.58}&-\\
            \midrule
            $\diamond$ w/o Feature F &92.88&1.51$\downarrow$&57.94&1.64$\downarrow$ \\
            $\diamond$ w/o Norm F &93.74&0.65$\downarrow$&56.23&3.35$\downarrow$ \\
            $\diamond$ w/o Conclusion F &91.03&3.36$\downarrow$&54.56&5.02$\downarrow$ \\
            $\diamond$ w/o Diversity R &92.15&2.24$\downarrow$&56.85&2.73$\downarrow$ \\
            \midrule
            $\diamond$ w/o All Parts &89.06&5.33$\downarrow$&53.02&6.56$\downarrow$ \\
            \midrule
            \midrule
            Mistral-7B &\textbf{97.66}&-&\textbf{66.98}&-\\
            \midrule
            $\diamond$ w/o Feature F &95.22&$\downarrow$2.44&65.04&$\downarrow$1.92 \\
            $\diamond$ w/o Norm F &95.98&$\downarrow$1.68&63.34&$\downarrow$3.62\\
            $\diamond$ w/o Conclusion F &93.61&$\downarrow$4.05&63.05&$\downarrow$3.91 \\
            $\diamond$ w/o Diversity R &95.54&$\downarrow$2.12&64.45&$\downarrow$2.51 \\
            \midrule
            $\diamond$ w/o All Parts &91.77&$\downarrow$5.89&61.91&$\downarrow$5.05\\
            \midrule
            \midrule
            Llama2-13B &\textbf{99.53}&-&\textbf{64.83}&-\\
            \midrule
            $\diamond$ w/o Feature F &96.27&$\downarrow$3.26&62.47&$\downarrow$2.36 \\
            $\diamond$ w/o Norm F &97.59&$\downarrow$1.94&61.34&$\downarrow$3.49\\
            $\diamond$ w/o Conclusion F &94.54&$\downarrow$4.99&61.07&$\downarrow$3.76 \\
            $\diamond$ w/o Diversity R &95.67&$\downarrow$3.86&62.33&$\downarrow$2.50 \\
            \midrule
            $\diamond$ w/o All Parts &93.01&$\downarrow$6.52&60.11&$\downarrow$4.72\\
            \midrule
		\bottomrule
	\end{tabular}
	\caption{Ablation study for MPT-7B, Llama2-7B, Mistral-7B and Llama2-13B. Macro F1-scores are exhibited, and \textbf{$\Delta$\textsubscript{All}} indicates score changes.}
    \label{tabs:ablation_all}
\end{table}

%% file: tabs/full_overall_performance.tex
\begin{table*}[!t]

    \renewcommand\arraystretch{1.06}
    \small
    \begin{center}           
    \begin{tabular}{m{2cm}|m{2cm}|m{0.95cm}<{\centering}m{0.95cm}<{\centering}m{0.95cm}<{\centering}|m{0.95cm}<{\centering}m{0.95cm}<{\centering}m{0.95cm}<{\centering}|m{0.95cm}<{\centering}m{0.95cm}<{\centering}}
            \toprule
            \multirow{2}{*}{\textbf{Method}} & \multirow{2}{*}{\textbf{Models}}& \multicolumn{3}{c|}{\textbf{Applicable}} &  \multicolumn{3}{c|}{\textbf{Not Applicable}}  & \multicolumn{2}{c}{\textbf{All}} \\
            && \textbf{Prec} & \textbf{Rec} & \textbf{F1}& \textbf{Prec} &  \textbf{Rec} & \textbf{F1}& \textbf{Acc} &  \textbf{Ma-F1}  \\
            \midrule
            \multirow{4}{*}{{\textbf{LLM API}}}
            &ChatGPT    &94.90&86.92&90.73&87.93&95.33&91.48&91.12&91.11\\
            &GPT-4      &97.17&96.26&96.71&96.30&97.20&96.74&\textbf{96.73}&\textbf{96.73}\\
            &ChatGPT~(MS)  &95.00&88.79&91.79&89.47&95.33&92.31&92.06&92.05\\
            &GPT-4~(MS)     &92.79&96.26&94.50&96.12&92.52&94.29&\underline{94.39}&\underline{94.39}\\
            \midrule
            \multirow{4}{*}{{\textbf{Zero-shot}}}
            &MPT-7B     &55.08&60.75&57.78&56.25&50.47&53.20&55.61&55.49\\ 
            &Llama2-7B  &65.22&98.13&78.36&96.23&47.66&63.75&72.90&71.05\\
            &Mistral-7B  &91.18&86.92&89.00&87.50&91.59&89.50&\underline{89.25}&\underline{89.25}\\
            &Llama2-13B   &98.89&83.18&90.36&85.48&99.07&91.77&\textbf{91.12}&\textbf{91.07}\\
            \midrule
            \multirow{4}{*}{{\textbf{Law Recitation}}}
            &MPT-7B      &44.21&39.25&41.58&45.38&50.47&47.79&44.86&44.69\\ 
            &Llama2-7B  &66.46&98.13&79.25&96.43&50.47&66.26&74.30&72.75\\
            &Mistral-7B  &88.89&82.24&85.44&83.48&89.72&86.49&\underline{85.98}&\underline{85.96}\\
            &Llama2-13B   &95.88&86.92&91.18&88.03&96.26&91.96&\textbf{91.59}&\textbf{91.57}\\
            \midrule
            \multirow{4}{*}{{\textbf{Direct Prompt}}}
            &MPT-7B    &100.00&27.10&42.65&57.84&100.00&73.29&63.55&57.97\\ 
            &Llama2-7B  &100.00&78.50&87.96&82.31&100.00&90.30&89.25&89.13\\
            &Mistral-7B  &100.00&90.65&95.10&91.45&100.00&95.54&\textbf{95.33}&\textbf{95.32}\\
            &Llama2-13B  &97.03&91.59&94.23&92.04&97.20&94.55&\underline{94.39}&\underline{94.39}\\
            \midrule
            \multirow{4}{*}{{\textbf{\papertitle}}}
            &MPT-7B     &77.46&51.40&61.80&63.64&85.05&72.80&68.22&67.30\\ 
            &Llama2-7B  &97.03&91.59&94.23&92.04&97.20&94.55&94.39&94.39\\
            &Mistral-7B  &100.00&95.33&97.61&95.54&100.00&97.72&\underline{97.66}&\underline{97.66}\\
            &Llama2-13B  &100.00&99.07&99.53&99.07&100.00&99.53&\textbf{99.53}&\textbf{99.53}\\
            \bottomrule
    \end{tabular}
    \end{center}
\caption{Performance of \papertitle and baselines under different settings across ``Applicable'' and ``Not Applicable'' categories. We \textbf{bold} the best results and \underline{underline} the second-best results in each setting. \textbf{MS} denotes the setting of employing multi-step instruction.}
\label{tabs:full_overall_performance_applicability}
\end{table*}

\begin{table*}[!t]

    \renewcommand\arraystretch{1.06}
    \small
    \begin{center}           
    \begin{tabular}{m{2cm}|m{2cm}|m{0.95cm}<{\centering}m{0.95cm}<{\centering}m{0.95cm}<{\centering}|m{0.95cm}<{\centering}m{0.95cm}<{\centering}m{0.95cm}<{\centering}|m{0.95cm}<{\centering}m{0.95cm}<{\centering}}
            \toprule
            \multirow{2}{*}{\textbf{Method}} & \multirow{2}{*}{\textbf{Models}}& \multicolumn{3}{c|}{\textbf{Permit}} &  \multicolumn{3}{c|}{\textbf{Forbid}}  & \multicolumn{2}{c}{\textbf{All}} \\
            && \textbf{Prec} & \textbf{Rec} & \textbf{F1}& \textbf{Prec} &  \textbf{Rec} & \textbf{F1}& \textbf{Acc} &  \textbf{Ma-F1}  \\
            \midrule
            \multirow{4}{*}{{\textbf{LLM API}}}
            &ChatGPT    &88.00&75.86&81.48&34.38&55.00&42.31&71.96&61.89\\
            &GPT-4      &87.21&86.21&86.71&42.86&45.00&43.90&\textbf{78.50}&\underline{65.30}\\
            &ChatGPT~(MS)   &86.59&81.61&84.02&36.00&45.00&40.00&\underline{74.77}&62.01\\
            &GPT-4~(MS)     &92.86&74.71&82.80&40.54&75.00&52.63&\underline{74.77}&\textbf{67.72}\\
            \midrule
            \multirow{4}{*}{{\textbf{Zero-shot}}}
            &MPT-7B     &77.78&48.28&59.57&15.09&40.00&21.92&46.73&40.75\\ 
            &Llama2-7B  &81.25&59.77&68.87&18.60&40.00&25.40&\underline{56.07}&47.14\\
            &Mistral-7B &94.74&41.38&57.60&26.09&90.00&40.45&50.47&\underline{49.02}\\
            &Llama2-13B  &86.76&67.82&76.13&28.21&55.00&37.29&\textbf{65.42}&\textbf{56.71}\\
            \midrule
            \multirow{4}{*}{{\textbf{Law Recitation}}}
            &MPT-7B       &70.37&43.68&53.90&7.55&20.00&10.96&39.25&32.43\\ 
            &Llama2-7B  &86.11&35.63&50.41&21.13&75.00&32.97&42.99&41.69\\
            &Mistral-7B  &78.46&58.62&67.11&14.29&30.00&19.35&\underline{53.27}&\underline{43.23}\\
            &Llama2-13B   &88.41&70.11&78.21&31.58&60.00&41.38&\textbf{68.22}&\textbf{59.79}\\
            \midrule
            \multirow{4}{*}{{\textbf{Direct Prompt}}}
            &MPT-7B     &85.92&70.11&77.22&27.78&50.00&35.71&\underline{66.36}&\underline{56.46}\\ 
            &Llama2-7B  &85.07&65.52&74.03&25.00&50.00&33.33&62.62&53.68\\
            &Mistral-7B  &97.44&43.68&60.32&27.94&95.00&43.18&53.27&51.75\\
            &Llama2-13B   &87.34&79.31&83.13&35.71&50.00&41.67&\textbf{73.83}&\textbf{62.40}\\
            \midrule
            \multirow{4}{*}{{\textbf{\papertitle}}}
            &MPT-7B     &86.49&73.56&79.50&30.30&50.00&37.74&69.16&58.62\\ 
            &Llama2-7B   &84.21&91.95&87.91&41.67&25.00&31.25&\textbf{79.44}&59.58\\
            &Mistral-7B   &90.67&78.16&83.95&40.62&65.00&50.00&75.70&\textbf{66.98}\\
            &Llama2-13B    &87.80&82.76&85.21&40.00&50.00&44.44&\underline{76.64}&\underline{64.83}\\
            \bottomrule
    \end{tabular}
    \end{center}
\caption{Performance of \papertitle and baselines under different settings across ``Permit'' and ``Forbid'' categories. We \textbf{bold} the best results and \underline{underline} the second-best results in each setting. \textbf{MS} denotes the multi-step instruction.}
\label{tabs:full_overall_performance_compliance}
\end{table*}

%% file: tabs/evluation_overall_performance_cot.tex
\begin{table*}[!t]

    \renewcommand\arraystretch{1.06}
    \small
    \begin{center}           
    \begin{tabular}{m{1.9cm}|m{1.9cm}|m{0.95cm}<{\centering}m{0.95cm}<{\centering}|m{0.95cm}<{\centering}m{0.95cm}<{\centering}|m{0.95cm}<{\centering}m{0.95cm}<{\centering}||m{0.95cm}<{\centering}m{0.95cm}<{\centering}}
            \toprule
            \multirow{2}{*}{\textbf{Task}}&\multirow{2}{*}{\textbf{Method}} & \multicolumn{2}{c|}{\textbf{MPT-7B}} &  \multicolumn{2}{c|}{\textbf{Llama2-7B}}  & \multicolumn{2}{c|}{\textbf{Mistral-7B}}&\multicolumn{2}{c}{\textbf{Llama2-13B}}   \\
            && {Acc} & {Ma-F1} & {Acc} &  {Ma-F1} & {Acc} &  {Ma-F1} & {Acc} &  {Ma-F1} \\
            \midrule
            \multirow{3}{*}{{\textbf{Applicability}}}
            &Zero-shot       &57.01&57.01&70.09&70.03&91.12&91.11&87.85&87.74\\ 
            &Law Recitation  &44.86&44.82&71.03&70.82&89.72&89.70&92.06&92.05\\
            &\textbf{\papertitle} &68.22&67.30&94.39&94.39&\underline{97.66}&\underline{97.66}&\textbf{99.53}&\textbf{99.53}\\
            
            \midrule
            \multirow{3}{*}{{\textbf{Compliance}}}
            &Zero-shot       &48.60&41.80&57.01&46.73&57.01&49.61&67.29&54.95\\ 
            &Law Recitation  &42.99&37.39&48.60&41.18&53.27&45.23&67.29&58.15\\
            &\textbf{\papertitle} &69.16&58.62&\textbf{79.44}&59.58&75.70&\textbf{66.98}&\underline{76.64}&\underline{64.83}\\
            \bottomrule
    \end{tabular}
    \end{center}
\caption{Performance of four LLMs with multi-step instruction setting, showing \textbf{Acc} and \textbf{Ma-F1} across both applicability and compliance tasks. We \textbf{bold} the best results and \underline{underline} the second-best results in each task.}
\vspace{-10pt}
    \label{tabs:overall_performance_cot}
\end{table*}

%% file: tabs/prompt_norm_classification.tex
\begin{table*}[t!]
\small
\centering
\begin{tabular}{p{2\columnwidth}}
\toprule
    Now you are a legal expert on HIPAA Privacy Rule that answers questions as simply as possible.\\
    Please read the following regulation {text}, and finish the following task.\\\\

    Q1: (Classification) Classify the regulation type of the following regulation. The regulation type is one of the following: ``Definition'', ``Permit'', ``Forbid'', ``Exception'', ``Requirement'', ``Permit and Exception'', ``Forbid and Exception'', ``Permit and Requirement'', ``Forbid and Requirement'', ``Permit and Exception and Requirement'', ``Forbid and Exception and Requirement'', ``Other''.\\

    Definition: The regulation defines a term or characteristic.\\
    Permit: The regulation permits certain actions regarding the flow of private information.\\
    Forbid: The regulation forbids certain actions regarding the flow of private information.\\
    Exception: The regulation defines an exception to a certain action about privacy information flow.\\
    Requirement: The regulation defines a requirement for privacy information flow.\\
    Other: The regulation is not in the above types.\\\\

    Q2: If the regulation type is ``Definition'', please annotate the name of the term or characteristic defined in the regulation.\\\\

    Q3: If the regulation type is ``Definition'', please annotate the definition of the term or characteristic defined in the regulation.\\\\

    Q4: If the regulation type contains ``Permit'', please annotate the action permitted in the regulation.\\\\

    Q5: If the regulation type contains ``Forbid'', please annotate the action forbidden in the regulation.\\\\

    Q6: If the regulation type contains ``Exception'', please annotate the exception defined in the regulation.\\\\

    Q7: If the regulation type contains ``Requirement'', please annotate the requirement defined in the regulation.\\\\

    Q8: If the regulation type is ``Other'', please give your own classification of the regulation type.\\
\bottomrule
\end{tabular}
\caption{Prompt of classifying norm types (\ie categories). GPT-4 is further instructed to provide details of each category.}
\label{tabs:norm_classfication_prompt}
\end{table*}

%% file: tabs/prompt_case_generation.tex
\begin{table*}[t!]

\small
\centering
\begin{tabular}{p{2\columnwidth}}
	\toprule
	Now you are a legal expert on HIPAA Privacy Rule that answers questions as simply as possible.
    \\
    Read the regulation: \{text\}.\\
    The regulation type is: \{type\}.\\
    \\
    Q1. Create a legal case only related to regulation {id} and the type {type}.
    The case must be a detailed story in plain text, spanning between 200 to 500 words, closely related to the regulation {id}.
    The story must include the following seven characteristics about the flow of private information: [Sender, Sender Role, Recipient, Recipient Role, Subject, Subject Role, Type].
    \\
    The ``Sender'' and ``Recipient'' ﬁelds indicate the sender and recipient of the message.
    The ``Sender Role'' and ``Recipient Role'' ﬁelds indicate the role of the sender and recipient (e.g., doctor, patient).
    The ``Subject'' and ``Subject Role'' ﬁeld identiﬁes the subject whose personal health information is contained in the message and the role of the subject.
    The ``Type'' ﬁeld deﬁnes what kind of information would be passed, such as name or location.
    Integrate these seven characteristics seamlessly into the story without explicitly listing them.

    Except for the seven characteristics, you can add the four optional characteristics [Purpose, In Reply To, Consented By, Belief] if mentioned in the regulation {id} or necessary.
    The ``Purpose'' ﬁeld indicates a reason the message is being sent, such as for medical treatment.
    The ``In Reply To'' ﬁeld was added to describe a disclosure where the message is sent as a response to some earlier message.
    The ``Consented By'' ﬁeld indicates which people have consented to the message disclosure.
    The ``Belief'' ﬁeld contains a collection of assertions about the current situation, such as whether this is a medical emergency, or whether the disclosure is (in the opinion of the sender) in the best interest of the health of the patient.
    Integrate these four characteristics seamlessly into the story without explicitly listing them.
    \\\\
    Q2: Based on the background created in Q3, list the eleven characteristics regarding the flow of private information (Mark as ``None'' if not exist)
    \\\\
    Q3: Please retrieve all the specific HIPAA regulation IDs that are the permission or prohibition description of the case. Please be as specific as possible to the sub-section id (e.g., 164.xxx).
    \\\\
    Q4: Please classify the relation between the case and the regulation {id} as one of the following: ``Permit'', ``Forbid'', ``Not Applicable''.
    \\\\
    Q5: Please classify the relation between the case and the HIPAA Privacy Rule as one of the following: ``Permit'', ``Forbid'', ``Not Applicable''.\\
    \bottomrule
\end{tabular}
\caption{Prompt of case generation. We guide GPT-4 to generate case backgrounds and other details through a series of questions.}
\label{tabs:case_generation_prompt}
\end{table*}

%% file: tabs/synthetic_case_example.tex
\begin{figure*}[t]
\begin{tcolorbox}[colback = cBlue_1!5, colframe = cBlue_6,  coltitle=white,fonttitle=\bfseries\small,fontupper=\small,fontlower=\small]
    \textbf{§ Seed Norm Id:} 164.502(j)(1)(i)\\
    \textbf{§ Seed Norm Content:}\\
    HIPAA: HIPAA Privacy Rule\\
    Part164: PART 164 — SECURITY AND PRIVACY\\
    Part164SubpartE: Subpart E—Privacy of Individually Identifiable Health Information\\
    164.502:§ 164.502 Uses and disclosures of protected health information: General rules.\\
    164.502(j):(j) Standard: Disclosures by whistleblowers and workforce member crime victims\\
    164.502(j)(1):(1) Disclosures by whistleblowers.  A covered entity is not considered to have violated the requirements of this subpart if a member of its workforce or a business associate discloses protected health information, provided that:\\
    164.502(j)(1)(i):(i) The workforce member or business associate believes in good faith that the covered entity has engaged in conduct that is unlawful or otherwise violates professional or clinical standards, or that the care, services, or conditions provided by the covered entity potentially endangers one or more patients, workers, or the public.\\
    \textbf{§ Case Background:} Jane, a nurse at Happy Valley Hospital, recently became aware of a situation where the hospital management was not following proper sterilization procedures for surgical instruments. She believed this could potentially endanger the health and safety of patients, other workers, and the public. Concerned about the situation, Jane decided to report the issue to the state health department. She gathered evidence, including protected health information about several affected patients, and sent it to a health department official, Tom. Jane believed in good faith that disclosing this information was necessary to report the hospital's misconduct. Tom, in his role at the health department, reviewed the information and initiated an investigation into the hospital's practices. In this case, the Sender is Jane, the Sender Role is a nurse, the Recipient is Tom, the Recipient Role is a health department official, and the Subject Role is patient. The Subject field includes the patients whose protected health information was shared in the disclosure. The Type of information shared includes protected health information related to the improper sterilization issue. The Purpose of the disclosure is to report the hospital's misconduct for violating professional and clinical standards.\\
    \textbf{§ Contextual Features:}\\Sender: Jane, Sender Role: Nurse, Recipient: Tom, Recipient Role: Health department official, \\Subject: Affected patients, Subject Role: Patients, Type: Protected health information, \\
    Purpose: Report hospital's misconduct, Belief: Disclosure is necessary to report hospital's misconduct\\
    \textbf{§ Compliance Conclusion:} Permit\\
\end{tcolorbox}
\caption{A synthetic case generated by \papertitle complies with HIPAA Privacy Rule.}
\label{tabs:synthetic_case_permit_example}
\end{figure*}

\begin{figure*}[t]
\begin{tcolorbox}[colback = cBlue_1!5, colframe = cBlue_6,  coltitle=white,fonttitle=\bfseries\small,fontupper=\small,fontlower=\small]
    \textbf{§ Seed Norm Id:} 164.502(a)(5)(ii)(B)(1)\\
    \textbf{§ Seed Norm Content:} \\
    HIPAA: HIPAA Privacy Rule\\
    Part164: PART 164 — SECURITY AND PRIVACY\\
    Part 164 SubpartE: Subpart E—Privacy of Individually Identifiable Health Information\\
    164.502:§ 164.502 Uses and disclosures of protected health information: General rules.\\
    164.502(a):(a) Standard.  A covered entity or business associate may not use or disclose protected health information, except as permitted or required by this subpart or by subpart C of part 160 of this subchapter.\\
    164.502(a)(5):(5) Prohibited uses and disclosures\\
    164.502(a)(5)(ii):(ii) Sale of protected health information:\\
    164.502(a)(5)(ii)(B):(B) For purposes of this paragraph, sale of protected health information means:\\
    164.502(a)(5)(ii)(B)(1):(1) Except as provided in paragraph (a)(5)(ii)(B)(2) of this section, disclosure of protected health information by a covered entity or business associate, if applicable, where the covered entity or business associate directly or indirectly receives remuneration from or on behalf of the recipient of the protected health information in exchange for the protected health information.\\
    \textbf{§ Case Background:} Jane Doe, a patient at XYZ Clinic, recently underwent a medical procedure. Dr. Smith, her treating physician at the clinic, documented her medical history, diagnosis, and treatment plan in her electronic health records. XYZ Clinic has a partnership with a pharmaceutical company, PharmaCorp, which is interested in obtaining patient data for research purposes. The clinic's administrator, without Jane's knowledge or consent, sends her protected health information (PHI) to PharmaCorp in exchange for financial remuneration.PharmaCorp's researcher, who receives Jane's PHI, analyzes it to develop new drugs and treatment plans. The researcher is aware that the information has been obtained in exchange for payment to the clinic. Meanwhile, Jane learns about this transaction and is upset that her PHI has been shared without her consent. She files a complaint with the Department of Health and Human Services (HHS).\\
    \textbf{§ Contextual Feature:} Sender: XYZ Clinic's administrator, Sender Role: Covered Entity, Recipient: PharmaCorp's researcher, Recipient Role: Business Associate, Subject: Jane Doe, Subject Role: Patient, Type: Protected Health Information (PHI), Purpose: Research\\
    \textbf{§ Compliance Conclusion:} Forbid\\
\end{tcolorbox}
\caption{A synthetic case generated by \papertitle does not comply with HIPAA Privacy Rule.}
\label{tabs:synthetic_case_forbid_example}
\end{figure*}

%% file: tabs/not_applicable_case_example.tex
\begin{figure*}[t]
\begin{tcolorbox}[colback = cBlue_1!5, colframe = cBlue_6,  coltitle=white,fonttitle=\bfseries\small,fontupper=\small,fontlower=\small]
    \textbf{§ Related Norm Id:} None\\
    \textbf{§ Case Background: }On July 10, 2006, the plaintiff filed a complaint against the United States pursuant to 26 U.S.C. § 7433 claiming that the Internal Revenue Service (“IRS”) wrongfully disclosed her tax return information to the public. Compl., Miller v. United States, No. 06-cv-01250 (D.D.C.), 6-12. On August 28, 2006, the plaintiff filed a second suit against the United States based on the same alleged misconduct, only this time complaining that the defendant had violated 26 U.S.C. § 7431. Compl. at ¶ 1. The § 7433 and § 7431 actions proceeded in parallel until the former was dismissed for failure to state a claim on July 19, 2007. In her remaining § 7431 suits, the plaintiff seeks damages for “substantial personal embarrassment, loss of goodwill, loss in credit .... and actual damages totaling \$65,000.” Am. Compl. at 19. The court permitted an amendment to the plaintiffs complaint on September 18, 2006, because no responsive pleadings had yet been filed. See Fed.R.CivP. 15(a). On November 6, 2006, the defendant moved to dismiss the plaintiffs amended complaint for lack of subject-matter jurisdiction and for failure to state a claim. The plaintiff did not file a response, but on January 22, 2007, she filed a motion to amend her complaint again. The court now turns to the merits of the government’s motion to dismiss the plaintiffs § 7431 claim and the plaintiffs motion to amend her complaint for a second time.\\
    \textbf{§ Contextual Feature:} Sender: Internal Revenue Service (IRS), Sender Role: Government agency, Recipient: Public, Recipient Role: Public, Subject: Plaintiff's tax return information, Subject Role: Plaintiff, Type: Tax return information\\
    \textbf{§ Applicability Conclusion:} Not Applicable\\
\end{tcolorbox}
\caption{A real court case sourced from CAP and is not relevant to HIPAA.}
\label{tabs:real_case_not_applicable_example}
\end{figure*}

%% file: tabs/prompt_real_case_process.tex
\begin{table*}[t!]

\small
\centering
\begin{tabular}{p{2\columnwidth}}
\toprule
    Now you are a legal expert on HIPAA Privacy Rule that answers questions as simply as possible.
    \\
    Read the case: \{case\}.\\
    \\
    Q1. If the case involves the flow of private information.
    Please annotate the eleven message characteristics [Sender, Sender Role, Recipient, Recipient Role, Subject, Subject Role, Type, Purpose, In Reply To, Consented By, Belief] about the flow of private information in the case as a list. If the characteristic does not exist, just fill in None.

    The ``Sender'' and ``Recipient'' ﬁelds indicate the sender and recipient of the message.
    The ``Sender Role'' and ``Recipient Role'' ﬁelds indicate the role of the sender and recipient (\eg doctor, patient).
    The ``Subject'' and ``Subject Role'' ﬁeld identiﬁes the subject whose personal health information is contained in the message and the role of the subject.
    The ``Type'' ﬁeld deﬁnes what kind of information would be passed, such as name or location.
    The ``Purpose'' ﬁeld indicates a reason the message is being sent, such as for medical treatment.
    The ``In Reply To'' ﬁeld was added to describe a disclosure where the message is sent as a response to some earlier message.
    The ``Consented By'' ﬁeld indicates which people have consented to the message disclosure.
    The ``Belief'' ﬁeld contains a collection of assertions about the current situation, such as whether this is a medical emergency, or whether the disclosure is (in the opinion of the sender) in the best interest of the health of the patient.
    \\\\
    Q2: Please retrieve all the specific HIPAA regulation IDs that are the permission or prohibition description of the case. Please be as specific as possible to the sub-section id (e.g., 164.xxx). If the regulations do not exist, just fill in None.
    \\\\
    Q3: Please classify the type of regulation(s). The regulation type is one of the following: ``Definition'', ``Permit'', ``Forbid'', ``Exception'', ``Requirement'', ``Permit and Exception'', ``Forbid and Exception'', ``Permit and Requirement'', ``Forbid and Requirement'', ``Permit and Exception and Requirement'', ``Forbid and Exception and Requirement'', ``Other''.
    \\\\
    Q4: Please classify the relation between the case and each regulation in Q3 as one of the following: ``Permit'', ``Forbid'', and ``Not Applicable''.
    \\\\
    Q5: A case may be associated with multiple regulations. If it is permitted by some regulations and not forbidden by any of the regulations, the case complies with HIPAA, answer ``Permit''. If it is not permitted by any of the regulations or forbidden by some regulations, the case violates HIPAA, answer ``Forbid''. Otherwise, if the case is not applicable to HIPAA, answer ``Not Applicable''. Please classify the relation between the flow of private information in the case and HIPAA as one of the following: ``Permit'', ``Forbid'', and ``Not Applicable''.
    \\\\
    Q6: With the eleven characteristics in Q2, restore the BACKGROUND story of the case, especially about the flow of private information.
    \\\\
    The case should not include any information about the regulation(s) in Q2 and the court decision.
    Make sure that the eleven characteristics are obviously included in the BACKGROUND story. The background must be a detailed story in plain text, spanning between 200 to 500 words.\\
\bottomrule
\end{tabular}
\caption{Prompt of parsing real court cases sourced from CAP. We guide GPT-4 through multiple questions to automatically extract HIPAA-related background stories for subsequent manual annotation.}
\label{tabs:real_case_process_prompt}
\end{table*}

%% file: tabs/real_case_example.tex
\begin{figure*}[t]
\begin{tcolorbox}[colback = cBlue_1!5, colframe = cBlue_6,  coltitle=white,fonttitle=\bfseries\small,fontupper=\small,fontlower=\small]
    \textbf{§ Related Norm Id:} 164.512(c)(1)\\
    \textbf{§ Case Background:} In Chester County, an 88-year-old woman, A.M., found herself at the center of a complex legal and medical situation. The county's Department of Aging Services received a confidential report suggesting that A.M., living alone, might be a victim of financial mismanagement or abuse. To investigate, the Department assigned Douglas Bernard as the investigator. Bernard, an experienced investigator, understood the gravity and sensitivity of his task. His role was to gather information on A.M.'s physical and mental health, as well as her financial situation, to determine if protective services were necessary. This included evaluating whether A.M. was capable of managing her affairs, especially in light of a diagnosis of dementia and hypertension. Initially, A.M. seemed cooperative. She agreed to meet her primary care physician, Dr. Priem and even gave consent for the release of her medical information. However, this cooperation was short-lived. Within days, she revoked the medical release and started to avoid any further engagement with the investigator. Bernard's attempts to visit her at home were met with silence; A.M. refused to answer her door. Despite these challenges, Bernard persisted in his investigation. He reached out to A.M.'s attorney, William McLaughlin, but found no help there either. A.M. had refused to grant McLaughlin consent to disclose any information to the investigator. In his report to the court, Bernard detailed his efforts and A.M.'s increasing uncooperativeness. He shared his concerns about A.M.'s mental health, citing her confusion and memory loss. He also expressed worry about her financial situation, noting the presence of large amounts of cash at her home and irregularities in her financial accounts. The court, faced with this information, had to make a critical decision. Was A.M. indeed at risk of abuse, neglect, or exploitation? Was she mentally and physically capable of managing her own affairs? And crucially, was the flow of her private information, without her explicit consent, justified in this context? A.M.'s case was a delicate balance between protecting her rights and ensuring her well-being.\\
    \textbf{§ Contextual Feature:}\\
    Sender: Douglas Bernard, Sender Role: Investigator, Recipient: Court, Recipient Role: Legal Authority, Subject: A.M. (the older adult), Subject Role: Subject of the Protective Services Investigation, Type: Medical and financial information, Purpose: To assess the need for protective services and to support the petition for involuntary intervention, In Reply To: The Department's requirement to assess A.M.'s situation under the Older Adults Protective Services Act, Consented By: None (A.M. revoked consent for the release of her medical information), Belief: The belief that A.M. is at imminent risk of abuse, neglect, exploitation, or abandonment and is incapacitated due to dementia\\
    \textbf{§ Compliance Conclusion:} Permit\\
\end{tcolorbox}
\caption{A real court case sourced from CAP complies with HIPAA Privacy Rule.}
\label{tabs:real_case_permit_example}
\end{figure*}

\begin{figure*}[t]
\begin{tcolorbox}[colback = cBlue_1!5, colframe = cBlue_6,  coltitle=white,fonttitle=\bfseries\small,fontupper=\small,fontlower=\small]
    \textbf{§ Related Norm Id:} 164.512(e)(1)\\
    \textbf{§ Case Background:} In December 1997, Richard Moss was involved in a traffic accident when Jennifer Amira rear-ended his vehicle. Moss, suffering from injuries, was taken to Northwest Community Hospital for immediate medical attention. Here, he received an examination, was fitted with a collar, given a prescription, and later released. Months later, in June 1998, Moss consulted Dr. Richard Moser, a neurological surgeon, for further evaluation. The encounter between Moss and Dr. Moser formed the basis of a subsequent medical opinion about the nature and cause of Moss's injuries. As the legal case progressed, defense counsel, representing Amira, sought to challenge Moss's claims about the extent and cause of his injuries. In February 2002, a discovery deposition of Dr. Moser was conducted, where he provided professional insights based on his examination and treatment of Moss. In a strategic move, the defense counsel sent a letter to Dr. Moser in April 2003, just before his evidence deposition. This letter contained a detailed narrative of the medical opinions expected to be disclosed at trial, including summaries of opinions from other treating physicians and those that Dr. Moser was expected to give following his discovery deposition. This letter outlined specific views about Moss's medical condition, its causes, and the necessity of surgery, which were crucial to the defense's argument. The letter did not have consent from Moss, the patient, and was part of a legal strategy to bolster the defense's case. The defense counsel believed this approach was necessary for case preparation and did not see it as a violation of any legal or ethical standards. However, this action led to a significant legal contention, as it was argued to be an inappropriate communication, potentially influencing the testimony of a treating physician. Moss's legal team saw this as a breach of the confidentiality and fiduciary relationship between a patient and his physician, raising concerns about the integrity of the legal process and the protection of private health information.\\
    \textbf{§ Contextual Feature:} Sender: Defense counsel, Sender Role: Attorney, Recipient: Dr. Richard Moser, Recipient Role: Doctor, Subject: Richard Moss, Subject Role: Patient, Type: Medical opinions and history, deposition excerpts, Purpose: To inform the physician about his expected opinions in the case, and potentially to influence the physician's future testimony, In Reply To: Plaintiff's supplemental opinion and the discovery deposition of Dr. Moser, Belief: The sender believed that this communication was necessary for case preparation and was not in violation of legal standards.\\
    \textbf{§ Compliance Conclusion:} Forbid\\
\end{tcolorbox}
\caption{A real court case sourced from CAP does not comply with HIPAA Privacy Rule.}
\label{tabs:real_case_forbid_example}
\end{figure*}